\documentclass[10pt,journal,compsoc]{IEEEtran}
\usepackage{url}
\usepackage{graphicx}
\usepackage{amsmath}
\usepackage[flushleft]{threeparttable}
\usepackage{tablefootnote}
\usepackage{booktabs}
\usepackage[export]{adjustbox}
\usepackage{array}
\newcolumntype{P}[1]{>{\centering\arraybackslash}p{#1}}
\newcolumntype{M}[1]{>{\centering\arraybackslash}m{#1}}
\usepackage{dblfloatfix}
\usepackage{algorithmic}
\usepackage{algorithm}
\usepackage{varwidth}

\ifCLASSOPTIONcompsoc
  \usepackage[nocompress]{cite}
\else
  \usepackage{cite}
\fi
\ifCLASSINFOpdf
\else
\fi
\ifCLASSOPTIONcompsoc
  \usepackage[caption=false,font=footnotesize,labelfont=sf,textfont=sf]{subfig}
\else
  \usepackage[caption=false,font=footnotesize]{subfig}
\fi
\begin{document}
%
\title{RaspiReader: Open Source Fingerprint Reader}
%
%
%
%

\author{Joshua~J.~Engelsma,
        Kai~Cao,
        and~Anil~K.~Jain,~\IEEEmembership{Life~Fellow,~IEEE}
\IEEEcompsocitemizethanks{\IEEEcompsocthanksitem J. J. Engelsma, K. Cao and A. K. Jain are with the Department of Computer Science and Engineering, Michigan State University, East Lansing, MI, 48824\protect\\
E-mail: \{engelsm7, kaicao, jain\}@cse.msu.edu
}
}

\IEEEtitleabstractindextext{%
\begin{abstract}
We open source an easy to assemble, spoof resistant, high resolution, optical fingerprint reader, called RaspiReader, using ubiquitous components. By using our open source STL files and software, RaspiReader can be built in under one hour for only US \$175. As such, RaspiReader provides the fingerprint research community a seamless and simple method for quickly prototyping new ideas involving fingerprint reader hardware. In particular, we posit that this open source fingerprint reader will facilitate the exploration of novel fingerprint spoof detection techniques involving both hardware and software. We demonstrate one such spoof detection technique by specially customizing RaspiReader with two cameras for fingerprint image acquisition. One camera provides high contrast, frustrated total internal reflection (FTIR) fingerprint images, and the other outputs direct images of the finger in contact with the platen. Using both of these image streams, we extract complementary information which, when fused together and used for spoof detection, results in marked performance improvement over previous methods relying only on grayscale FTIR images provided by COTS optical readers. Finally, fingerprint matching experiments between images acquired from the FTIR output of RaspiReader and images acquired from a COTS reader verify the interoperability of the RaspiReader with existing COTS optical readers.
\end{abstract}

\begin{IEEEkeywords}
Raspberry Pi, Frustrated Total Internal Reflection (FTIR), Open Source Fingerprint Readers, Presentation Attack Detection, Spoof Detection, Interoperability
\end{IEEEkeywords}}

\maketitle

\IEEEdisplaynontitleabstractindextext

%
\IEEEpeerreviewmaketitle

\IEEEraisesectionheading{\section{Introduction}\label{sec:introduction}}

%
%
%
%
\IEEEPARstart{O}{ne} of the major challenges facing biometric technology today is the growing threat of presentation attacks\footnote{In ISO standard IEC 30107-1:2016(E), presentation attacks are defined as the {\it ``presentation to the biometric data capture subsystem with the goal of interfering with the operation of the biometric system"}\cite{iso}.}~\cite{odin}. The most common type of presentation attack (referred to as spoofing) occurs when a {\it hacker} intentionally assumes the identity of unsuspecting individuals, called {\it victims} here, through stealing their fingerprints, fabricating spoofs with the stolen fingerprints, and maliciously attacking fingerprint recognition systems with the spoofs into identifying the hacker as the victim\footnote{Presentation attacks can also occur when (i) two individuals are in collusion or (ii) an individual obfuscates his or her own fingerprints to avoid recognition \cite{spoofs_survey}. However, in this paper our specific aim is to stop fingerprint spoofing presentation attacks.}  \cite{spoofs_survey, handbook, altered_yoon, gummy}.

\begin{figure}[t]
\begin{center}
\includegraphics[scale=0.18]{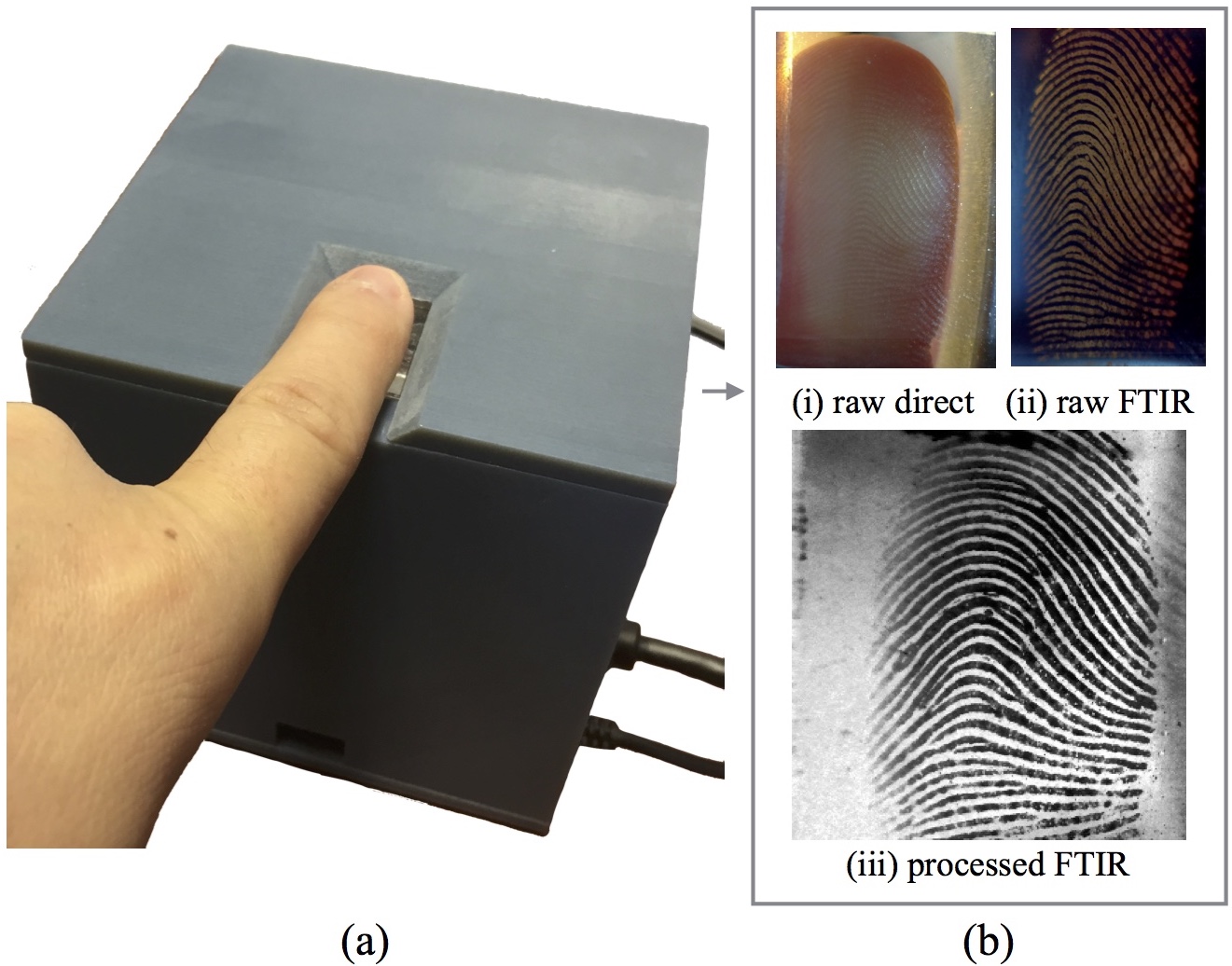}
\caption{Prototype of RaspiReader: two fingerprint images (b, (i)) and (b, (ii)) of the input finger (a) are captured. The raw direct image (b, (i)) and the raw, high contrast FTIR image (b, (ii)) both contain useful information for spoof detection. Following the use of (b, (ii)) for spoof detection, image calibration and processing are performed on the raw FTIR image to output a high quality, 500 ppi fingerprint for matching (b, (iii)). The dimensions of the RaspiReader shown in (a) are 100 {\it mm} x 100 {\it mm} x 105 {\it mm} (about the size of a 4 inch cube).
}
\label{fig:intro_fig}
\end{center}
\end{figure} 

\begin{figure*}[t]
\begin{center}
\includegraphics[scale=0.25]{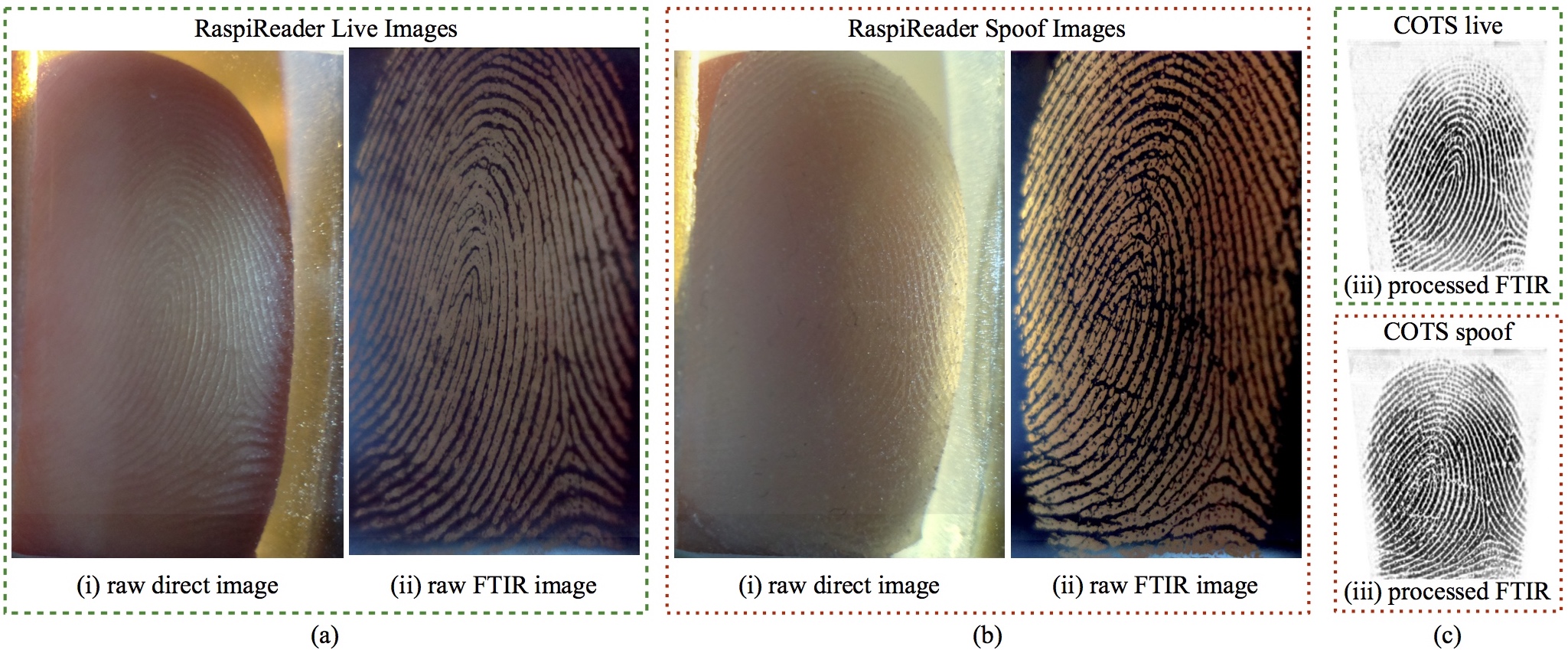}
\caption{Fingerprint images acquired using the RaspiReader. Images in (a) were collected from a live finger. Images in (b) were collected from a spoof finger. Using features extracted from both raw image outputs ((i), direct) and ((ii), FTIR) of the RaspiReader, our spoof detectors are better able to discriminate between live fingers and spoof fingers. The raw FTIR image output of the RaspiReader (ii) can be post processed (after spoof detection) to output images suitable for fingerprint matching. Images in (c) were acquired from the same live finger (a) and spoof finger (b) on a commercial off-the-shelf (COTS) 500 ppi optical reader. The close similarity between the two images in (c) qualitatively illustrates why current spoof detectors are limited by the low information content, processed fingerprint images (c, (iii)) output by COTS readers. }
\label{fig:fingers}
\vspace{-1.0em}
\end{center}
\end{figure*}

The need to prevent spoof attacks is becoming increasingly urgent due to the monumental costs and loss of user privacy associated with spoofed systems. Consider for example India's Aadhaar program which (i) provides benefits and services to an ever growing population of over 1.2 billion residents through fingerprint and/or iris recognition \cite{india1, india2} and (ii) facilitates electronic financial transactions through the Unified Payments Interface (UPI) \cite{india3}.  Failure to detect spoof attacks in the Aadhaar system could cause the disruption of a commerce system affecting untold numbers of people. Also consider the United States Office of Biometric Identity Management (US OBIM) which is responsible for supporting the Department of Homeland Security (DHS) with biometric identification services specifically aimed at preventing people who pose security risks to the United States from entering the country \cite{obim}. Failure to detect spoofs on systems deployed by OBIM could result in a deadly terrorist attack\footnote{In 2012, a journalist successfully demonstrated that the Hong Kong-China border control system could be easily spoofed \cite{hongkongchina}.}. Finally, almost all of us are actively carrying fingerprint recognition systems embedded within our personal smart devices. Failure to detect spoof attacks on smartphones \cite{caophone} could compromise emails, banking information, social media content, personal photos and a plethora of other confidential information.

In an effort to mitigate the costs associated with spoof attacks, a number of spoof detection techniques involving both hardware and software have been proposed in the literature. Special hardware embedded in fingerprint readers\footnote{Several fingerprint vendors have developed hardware spoof detection solutions by employing multispectral imaging, infrared imaging (useful for sub-dermal finger analysis), and pulse capture to distinguish live fingers from spoof fingers \cite{com1, com3}.} enables capture of features such as heartbeat, thermal output, blood flow, odor, and sub-dermal finger characteristics useful for distinguishing a live finger from a spoof \cite{blood_flow, odor, com1, rowe1, oct, spoofs_survey, hardware_issues, burned, schuckers}. Spoof detection methods in software are based on extracting textural \cite{texture0, texture1, texture2, texture3, texture4}, anatomical \cite{pores}, and physiological \cite{perspiration1, perspiration2} features from processed\footnote{Raw fingerprint images are ``processed" (such as RGB to grayscale conversion, contrast enhancement, and scaling) by COTS readers to boost matching performance. However, useful spoof detection information (such as color and/or minute textural abberations) is lost during this processing.} fingerprint images which are used in conjunction with a classifier such as Support Vector Machines (SVM). Alternatively, a Convolutional Neural Network (CNN) can be trained to distinguish a live finger from a spoof \cite{CNN1, CNN2, tarang}. 

\begin{figure*}[t]
\begin{center}
\includegraphics[scale=0.25]{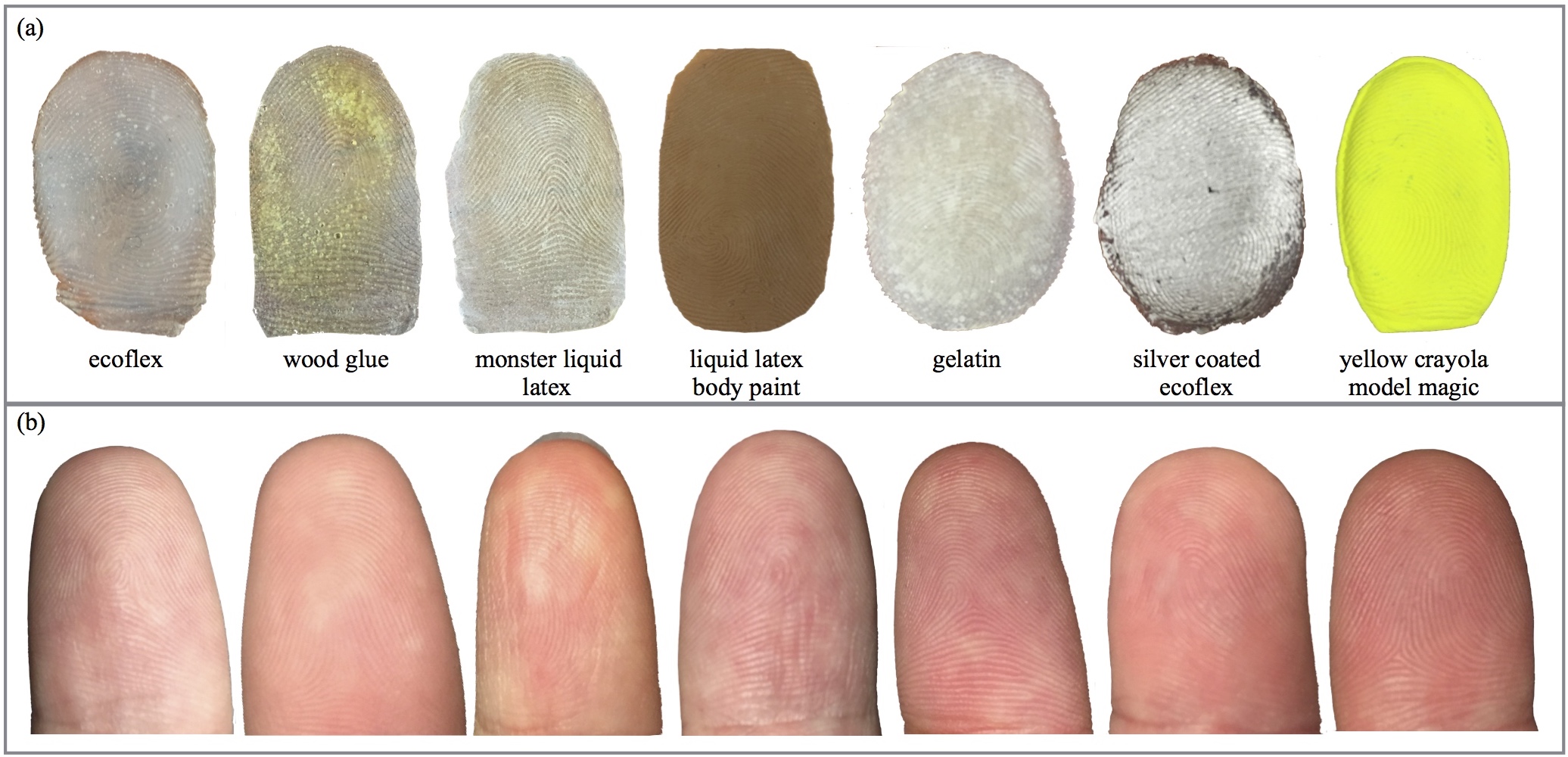}
\caption{Example spoof fingers and live fingers in our database. (a) Spoof fingers and (b) live fingers used to acquire both spoof fingerprint impressions and live fingerprint impressions for conducting the experiments reported in this paper. The spoofs in (a) and the live fingers in (b) are not in 1-to-1 correspondence.}
\label{fig:spoofs}
\vspace{-1.0em}
\end{center}
\end{figure*}

While existing hardware and software spoof detection schemes provide a reasonable starting point for solving the spoof detection problem, current solutions have a plethora of shortcomings. As noted in \cite{hardware_issues, burned, schuckers} most hardware based approaches can be easily bypassed by developing very thin spoofs (Fig. \ref{fig:spoofs} (a)), since heartbeat, thermal output, and blood flow can still be read from the live human skin behind the thin spoof. Additionally, some of the characteristics (such as odor and heartbeat) acquired by the hardware vary tremendously amongst different human subjects, making it very difficult to build an adequate model representative of all live subjects \cite{hardware_issues, burned}.

Current spoof detection software solutions have their own limitations. Although the LivDet 2015 competition reported state-of-the-art spoof detection software to have an average accuracy of 95.51\% \cite{2015}, the spoof detection performance at desired operating points such as False Detect Rate (FDR) of 0.1\% was not reported, and very limited evaluation was performed to determine the effects of testing spoof detectors with spoofs fabricated from materials not seen during training (cross-material evaluation). In the limited cross material evaluation that was performed, the rate of spoofs correctly classified as spoofs was shown to drop from 96.57\% to 94.20\% \cite{2015}. While this slight drop in accuracy seems promising, without knowing the performance at field conditions, namely False Detect Rate (FDR)\footnote{The required operating point for the ODIN program supporting this research is FDR = 0.2\%} of 0.1\% on a larger collection of unknown materials, the reported levels of total accuracy should be accepted with caution. Chugh et al. \cite{tarang} pushed state-of-the-art fingerprint spoof detection performance on the LivDet 2015 dataset from 95.51\% average accuracy to 98.61\% average accuracy using a CNN trained on patches around minutiae points, but they also demonstrated that performance at strict operating points dropped significantly in some experiments. For example, Chugh et al. reported an average accuracy on the LivDet 2011 dataset of 97.41\%, however, at a FDR of 1.0\%, the TDR was only 90.32\%, indicating that current state-of-the-art spoof detection systems leave room for improvement at desired operating points. Finally, several other studies have reported up to a three-fold increase in error when testing spoof detectors on unknown material types \cite{inter1,inter2,2011}. 

Because of the less than desired performance of spoof detection software to adapt to spoofs fabricated from unseen materials, studies in \cite{open1}, \cite{open2}, and \cite{open3} developed open-set recognition classifiers to better detect spoofs fabricated with novel material types. However, while these classifiers are able to generalize to spoofs made with new materials better than closed-set recognition algorithms, their overall accuracy (approx. 85\% - 90\%) still does not meet the desired performance for field deployments. 

Given the limitations of state-of-the-art fingerprint spoof detection (both in hardware and software), it is evident that much work remains to be done in developing robust and generalizable spoof detection solutions. We posit that one of the biggest limitations facing the most successful spoof detection solutions to date (such as use of textural features \cite{2011} and CNNs \cite{CNN1, CNN2, tarang}), is the processed COTS fingerprint reader images used to train spoof detectors. In particular, because COTS fingerprint readers output fingerprint images which have undergone a number of image processing operations (in an effort to achieve high matching performance), they are not optimal for fingerprint spoof detection, since valuable information such as color and textural aberrations is lost during the image processing operations. By removing color and minute textural details from the raw fingerprint images, spoof fingerprint impressions and live fingerprint impressions (acquired on COTS optical readers) appear very similar (Fig. \ref{fig:fingers} (c)), even when the physical live/spoof fingers used to collect the respective fingerprint impressions appear very different (Fig. \ref{fig:spoofs}). 

\begin{figure}[h]
\begin{center}
\includegraphics[scale=0.21]{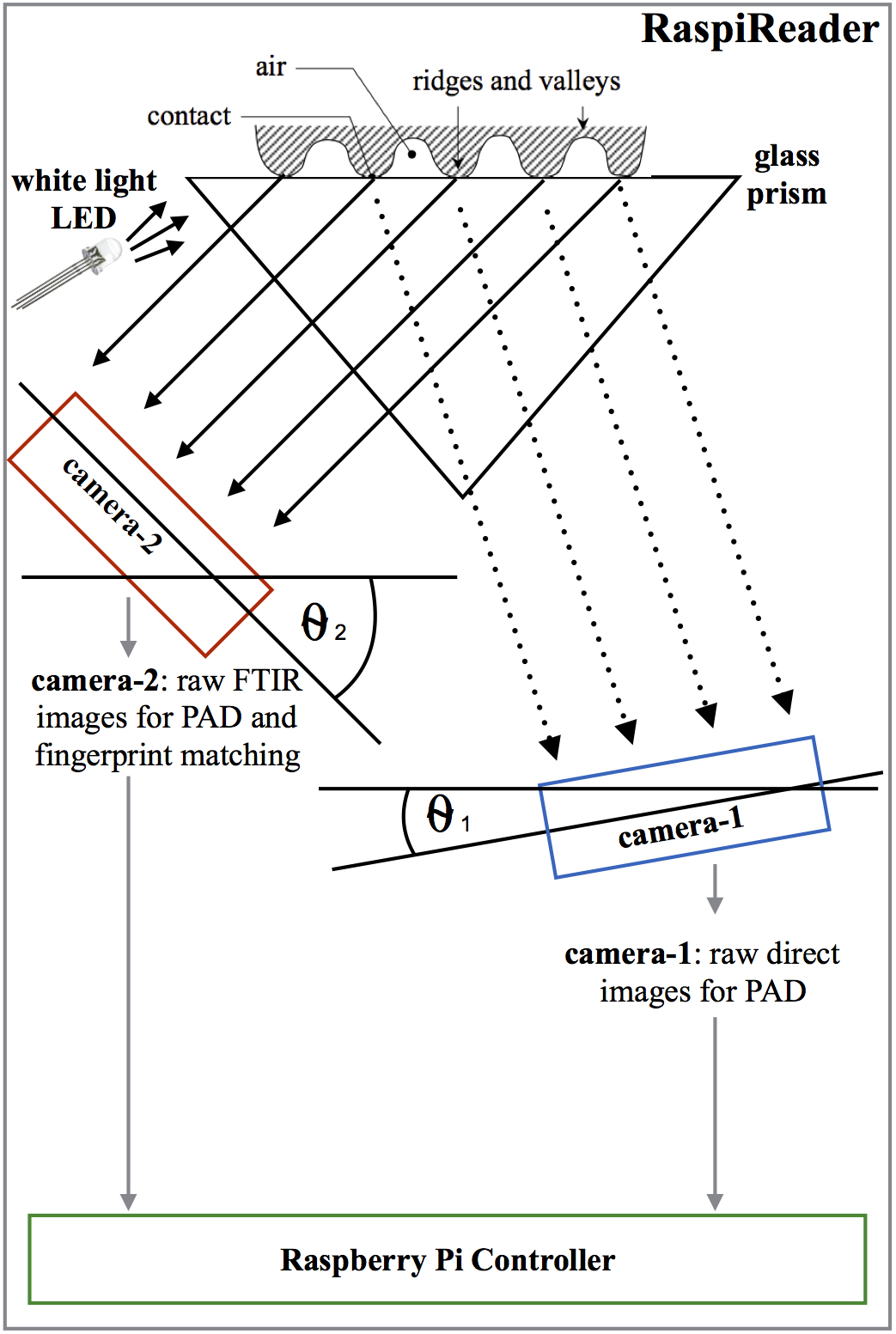}
\caption{Schematic illustrating RaspiReader functionality. Incoming white light from three LEDs enters the prism. Camera 2 receives light rays reflected from the fingerprint ridges only (light rays are not reflected back from the fingerprint valleys due to total internal reflection (TIR)). This image from Camera 2, with high contrast between ridges and valleys can be used for both spoof detection and fingerprint matching. Camera 1 receives light rays reflected from both the ridges and valleys. This image from Camera 1 provides complementary information for spoof detection.}
\label{fig:schematic}
\vspace{-1.0em}
\end{center}
\end{figure}

This limitation inherent to many existing spoof detection solutions motivated us to develop a custom, optical fingerprint reader, called RaspiReader, with the capability to output 2 raw images (from 2 different cameras) for spoof detection. By mounting two cameras at appropriate angles to a glass prism (Fig. \ref{fig:schematic}), one camera is able to capture high contrast FTIR fingerprint images (useful for both fingerprint spoof detection and fingerprint matching) (Fig. \ref{fig:fingers} (ii)), while the other camera captures direct images of the finger skin in contact with the platen (useful for fingerprint spoof detection) (Fig. \ref{fig:fingers} (i)). Both images of the RaspiReader visually differentiate between live fingers and spoof fingers much more than the processed fingerprint images output by COTS fingerprint readers (Fig. \ref{fig:fingers} (c)). 

RaspiReader's two camera approach is similar to that which was prescribed by Rowe et al. in \cite{com1, rowe1} where both an FTIR image and a direct view image were acquired using different wavelength LEDs, however, the commercial products developed around the ideas in \cite{com1, rowe1} act as a proprietary black box outputting only a single processed composite image of a collection of raw image frames captured under various wavelengths. As such, fingerprint researchers cannot implement new spoof detection schemes on the individual raw frames captured by the reader. Furthermore, unlike the patented ideas in \cite{com1}, RaspiReader is built with ubiquitous components and open source software packages, enabling fingerprint researchers to very easily prototype their own RaspiReader, further customize it with new spoof detection hardware, and gain direct access to the raw images captured by the reader. In short, the low cost (\$175 USD) and easy to implement (1 hour build time) RaspiReader is a truly unique concept which we posit will push the boundaries of state-of-the-art fingerprint spoof detection, by facilitating spoof detection schemes which use both hardware and software. 

Experiments demonstrate that by utilizing the two cameras of RaspiReader, we are able to significantly boost the performance of state-of-the-art spoof detectors previously trained on COTS grayscale images (both on known-material and cross-material testing scenarios). In particular, because both image outputs of the RaspiReader are raw and contain useful color information, we can extract discriminative and complementary information from each of the image outputs. By fusing this complementary information (at a feature level or score level) the performance of spoof detectors is significantly higher than when features are extracted from COTS grayscale images. 

Finally, by calibrating and processing the FTIR image output of the RaspiReader (post spoof detection), we demonstrate that RaspiReader is not only interoperable with existing COTS optical readers but is also capable of achieving state-of-the-art fingerprint matching accuracy. Note that interoperability with existing COTS readers is absolutely vital in any new hardware based spoof detection solution as it makes the spoof resistant device compatible (in terms of matching) with legacy fingerprint databases\footnote{Interoperability with existing COTS readers is a strict requirement of the IARPA ODIN program supporting this research \cite{odin}.}. Furthermore, by making the RaspiReader compatible with existing COTS readers, we further extend the utility of RaspiReader beyond spoof detection. In particular, RaspiReader is not only useful for providing direct access to multiple raw images for spoof detection; it also provides researchers in fingerprint matching the easy ability to fine tune (resolution and processing) the images being output by the fingerprint reader. In any imaging system, the recognition performance depends on the quality of the image output by the sensor. This is particularly true of fingerprint recognition systems. As shown in the NIST FpVTE 2013 \cite{nist} results, the single most important factor responsible for degrading fingerprint recognition performance is the fingerprint image quality. However, most fingerprint researchers have no control over the quality of the fingerprint images being used to develop fingerprint recognition algorithms since they must rely on blackbox COTS fingerprint readers. RaspiReader changes this by providing fingerprint matching algorithm designers an easy method for prototyping their own fingerprint reader and optimizing fingerprint image quality and fingerprint matching algorithms jointly in an effort to further improve fingerprint recognition performance.

\newcommand{\specialcell}[2][c]{%
  \begin{tabular}[#1]{@{}c@{}}#2\end{tabular}}
  \newcommand{\tabitem}{~~\llap{\textbullet}~~}

\begin{table*}[t]
\begin{center}
\begin{threeparttable}
\label{tab:parts}
\resizebox{\textwidth}{!}{%
\begin{tabular}{ |c||M{5.5cm}|c|c|}
 \multicolumn{4}{c}{{\tiny Table 1: Primary Components Used to Construct RaspiReader. Total Cost is \$175.20 (as of December 12, 2017)}} \\
 \hline
{\tiny Component Image} & {\tiny Name and Description} & {\tiny Quantity} & {\tiny Cost (USD)\tnote{1}}\\
\hline
 \includegraphics[valign=m,scale=0.10]{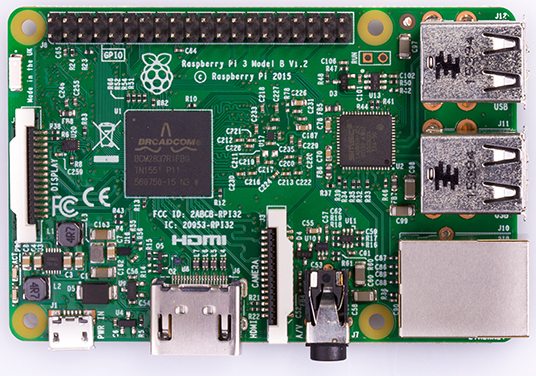} & {\tiny \textbf{Raspberry Pi 3B:}  A single board computer (SBC) with 1.2 GHz 64-bit quad-core CPU, 1 GB RAM, MicroSDHC storage, and Broadcom VideoCore IV Graphic card} & {\tiny1} & {\tiny \$38.27}\\
 \hline
 \includegraphics[valign=m,scale=0.075]{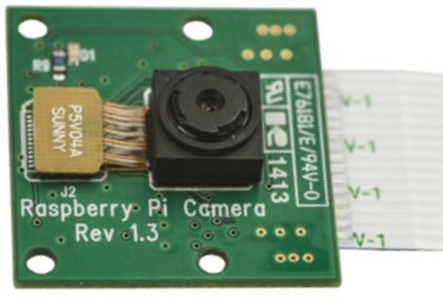}  & {\tiny \textbf{Raspberry Pi Camera Module V1:} A 5.0 megapixel, 30 frames per second, fixed focal length camera} & {\tiny 2} & {\tiny \$13.49}\\
  \hline
 \includegraphics[valign=m,scale=0.20]{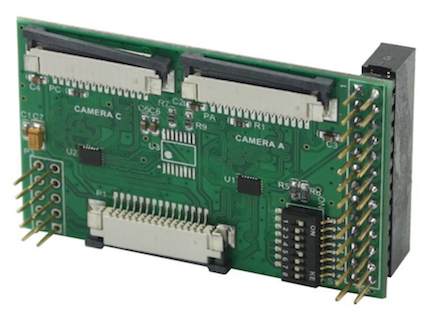} & {\tiny \textbf{Multi-Camera Adapter:} Splits Raspberry Pi camera slot into two slots, enabling connection of two cameras} & {\tiny 1} & {\tiny \$49.99}\\
  \hline
 \includegraphics[valign=m,scale=0.10]{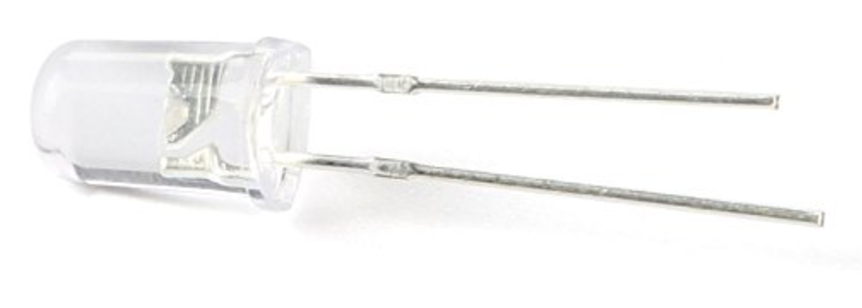}  & {\tiny \textbf{LEDs:} white light, 5 mm, 1 watt} & {\tiny3} & {\tiny \$0.10}\\
  \hline
\includegraphics[valign=m,scale=0.15]{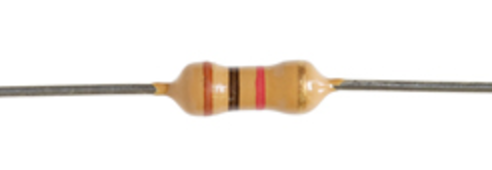} & {\tiny \textbf{Resistors:} 1 k$\Omega$} & {\tiny3} & {\tiny \$5.16}\\
  \hline
\includegraphics[valign=m,scale=0.15]{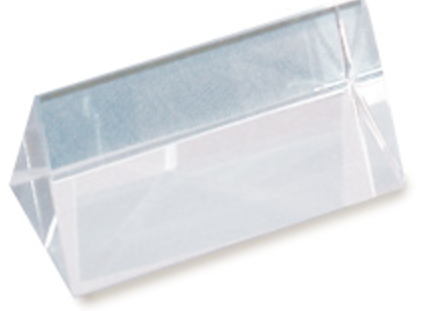}  & {\tiny \textbf{Right Angle Prism:\tnote{2}}~ 25 mm leg, 35.4 mm hypotenuse} & {\tiny 1} & {\tiny \$54.50}\\
\hline
 
\end{tabular}}
\begin{tablenotes}
\item[1] All items except the glass prism were purchased for the listed prices on Amazon.com
\item[2] The glass prism was purchased from ThorLabs \cite{thor}.
\end{tablenotes}
\end{threeparttable}
\end{center}
\vspace{-1.0em}
\end{table*} 

In summary, our work on RaspiReader removes the mystery of designing and understanding the internals of a fingerprint reader. Using the open-source fabrication process of this fingerprint reader, any fingerprint algorithm designer can quickly and affordably construct his or her own reader with the capabilities (spoof detection and matching image quality) necessary to meet their application requirements.

More concisely, the contributions of this research are:

\begin{itemize}
\item An open source, easy to assemble, cost effective fingerprint reader, called RaspiReader, capable of producing fingerprint images useful for spoof detection and that are of high quality and resolution (1,500 ppi - 3,300 ppi native resolution) for fingerprint matching. The custom RaspiReader can be easily modified to facilitate spoof detection and fingerprint matching studies.

\item A customized fingerprint reader with two cameras for image acquisition rather than a single camera. Use of two cameras enables robust fingerprint spoof detection, since we can extract features from two complementary, information rich images instead of processed grayscale images output by traditional COTS optical fingerprint readers. 
  
  \item A significant boost in spoof detection performance (both known-material and seven cross-material testing scenarios) using current state-of-the-art software based spoof detection methods in conjunction with RaspiReader images as opposed to COTS optical grayscale images. Spoofs of seven materials were used in both known-material and cross-material testing scenarios.

\item Demonstrated matching interoperability of RaspiReader with a COTS optical fingerprint reader. Since RaspiReader is shown to be interoperable with COTS readers, it could immediately be deployed in the real world since interoperability makes the device compatible with legacy fingerprint databases.
  
\end{itemize}

\section{RaspiReader Construction and Calibration}

In this section, the construction of the RaspiReader using ubiquitous, off-the-shelf components (Table 1) is explained. In particular, the main steps involved in constructing RaspiReader consist of (i) properly mounting cameras (angle and position) with respect to a glass prism, (ii) fabricating a plastic case to house the hardware components, (iii) assembling the cameras and hardware within the plastic case, and (iv) writing software to capture fingerprint images with the assembled hardware. Each of these steps is described in more detail in the following subsections. Finally, we provide the steps for calibrating and processing the raw FTIR fingerprint images of the RaspiReader for fingerprint matching.

\subsection{Camera Placement}

The most important step in constructing RaspiReader is the placement (angle and position) of the two cameras capturing fingerprint images. In particular, to collect an FTIR image of a fingerprint, a camera needs to be mounted at an angle greater than the critical angle, and to collect a direct view image, a camera needs to be mounted an an angle less than the critical angle (both with respect to the platen). Here, the critical angle is defined as the angle at which total internal reflection occurs when light passes from a medium with an index of refraction $n_1$ to another medium with index of refraction $n_2$ (Eq. \ref{eq:1}):

\begin{equation} \label{eq:1}
\theta_c=arcsin(\frac{n_2}{n_1})
\end{equation}

In the case of fingerprint sensing, the first medium is glass which has an index of refraction $n_1 = 1.5$, and the second medium is air which has an index of refraction of $n_2 = 1.0$ leading to a critical angle ($\theta_c$) of $41.8^\circ$. Therefore, as shown in (Fig. \ref{fig:schematic}), we mount the direct view camera ($camera_1$) at an angle of $\theta_1=10^\circ$ and we mount the FTIR camera ($camera_2$) at an angle of $\theta_2=45^\circ$.

With respect to the position of each camera lens to the glass prism, there is a tradeoff between resolution and fingerprint area to consider. As the camera is moved closer to the prism, the fingerprint image resolution (pixels per inch) is increased. However, if the cameras are too close to the platen, only part of the fingerprint image is within the field of view (FOV). In constructing RaspiReader, we wanted to maximize the fingerprint image resolution, while still capturing the entire fingerprint image within the FOV. We experimentally determined that at a distance of 23 {\it mm} from the prism, the cameras would capture the entire fingerprint area. At closer distances, part of the fingerprint image would start to be outside the FOV. As a final step in camera placement, the focal length of the Raspicams (cameras used in RaspiReader) must be increased so that the camera will focus on the nearby glass prism (the default focus-length of the Raspicams is 1 meter; much greater than the 23 {\it mm} distant prism). By default, the Raspicams have a fixed-focal length of 3.6 {\it mm}. However, by rotating the Raspicam lens $652.5^\circ$ counterclockwise (for the FTIR imaging camera) and $405^\circ$ counterclockwise (for the direct imaging camera), the focal length can be slightly increased to bring the nearby fingerprint images into focus. 

\subsection{Case Fabrication}

\begin{figure}[h]
\begin{center}
\includegraphics[scale=0.175]{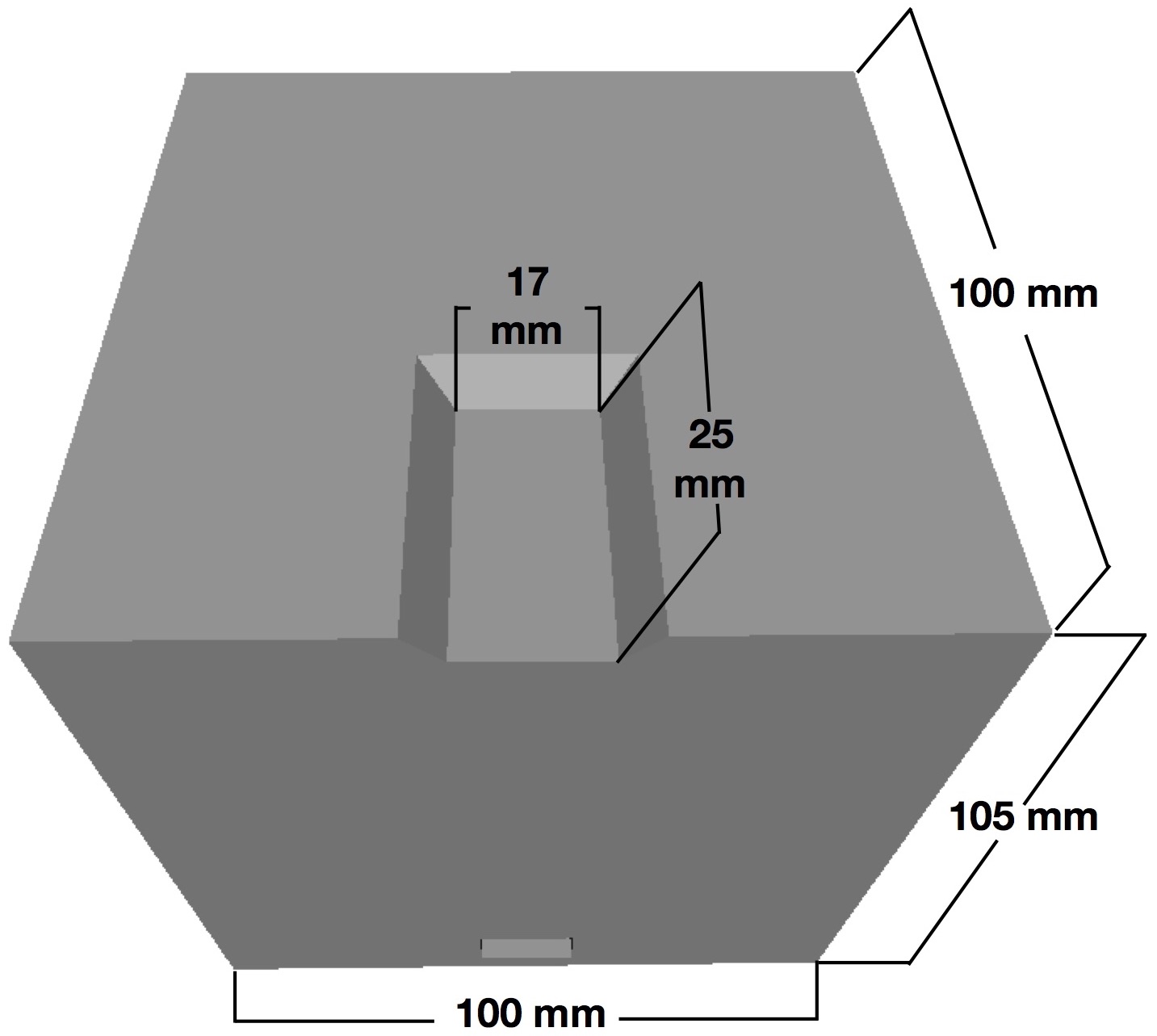}
\caption{Electronic CAD model of the RaspiReader case. The dimensions here were provided to a 3D printer for fabricating the prototype.}
\label{fig:case}
\vspace{-1.0em}
\end{center}
\end{figure}

\begin{figure}[h]
\begin{center}
\includegraphics[scale=0.5]{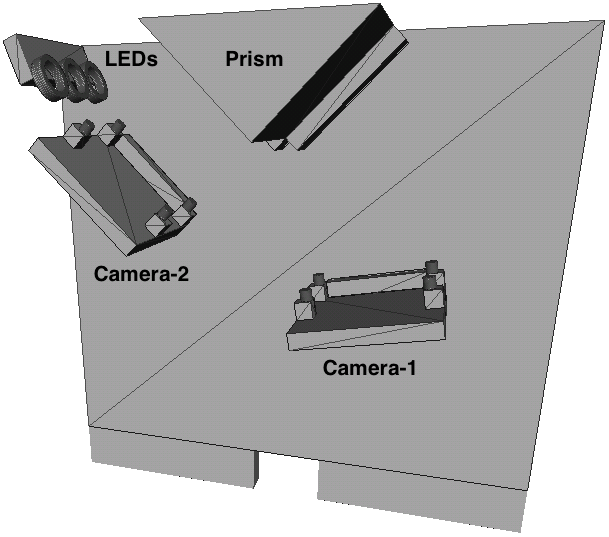}
\caption{Inside view of the RaspiReader case. The camera and LED mounts are positioned at the necessary angles and distance to the glass prism, making the reproduction of RaspiReader as simple as 3D printing the open-sourced STL files.}
\label{fig:mounts}
\vspace{-1.0em}
\end{center}
\end{figure}

After determining the angle and position of both cameras, an outer casing (Fig. \ref{fig:case}) accommodating these positions is electronically modeled using Meshlab \cite{meshlab} and subsequently 3D printed on a high resolution 3D printer (Stratasys Objet350 Connex)\footnote{We are currently investigating alternative case manufacturing methods such as CNC milling.}. To make the fabrication process easily reproducible, the camera mounts and light source mounts are modeled in place on the front part of the fingerprint reader case (Fig. \ref{fig:mounts}). As such, one only needs to 3D print the open-source STL files and clip the LEDs and Raspicams to their respective mounts (Fig. \ref{fig:mounts}) in order to quickly build their own RaspiReader replica.

\subsection{Image Acquisition Hardware and Software}

The backbone of the RaspiReader is the popular Raspberry Pi 3B single board computer, which enables easy interfacing with GPIO pins (for controlling LEDs) and image acquisition (with its standard camera and camera connection port). Because the Raspberry Pi only has a single camera connection port, a camera port multiplexer is used to enable the use of multiple cameras on a single Pi \cite{multi}. Using the Raspberry Pi GPIO pins, the code available in \cite{multi}, and the camera multiplexer, one can easily extend the Raspberry Pi to use multiple cameras.

After assembling the camera port multiplexer to the Pi (with two Raspicams), wiring 3 LEDs to the Raspberry Pi GPIO pins, and attaching the Raspicams and LEDs to the 3D printed casing mounts (Fig. \ref{fig:mounts}), open source python libraries \cite{multi} can be used to illuminate the glass prism and subsequently acquire two images (Fig. \ref{fig:fingers} (a)) from the fingerprint reader (one raw FTIR fingerprint image and another raw direct fingerprint image). 

\subsection{Fingerprint Image Processing}

In order for the RaspiReader to be used for spoof detection, it must also demonstrate the ability to output high quality fingerprint images suitable for fingerprint matching. As previously mentioned, the RaspiReader performs spoof detection on non-processed, raw fingerprint images. While these raw images are shown to provide discriminatory information for spoof detection, they need to be made compatible with processed images output by other COTS fingerprint readers. Therefore, after spoof detection, the RaspiReader performs image processing operations on the raw high contrast, FTIR image frames in order to output high fidelity images compatible with COTS optical fingerprint readers. 

\begin{figure*}[t]
\begin{center}
\includegraphics[scale=.325]{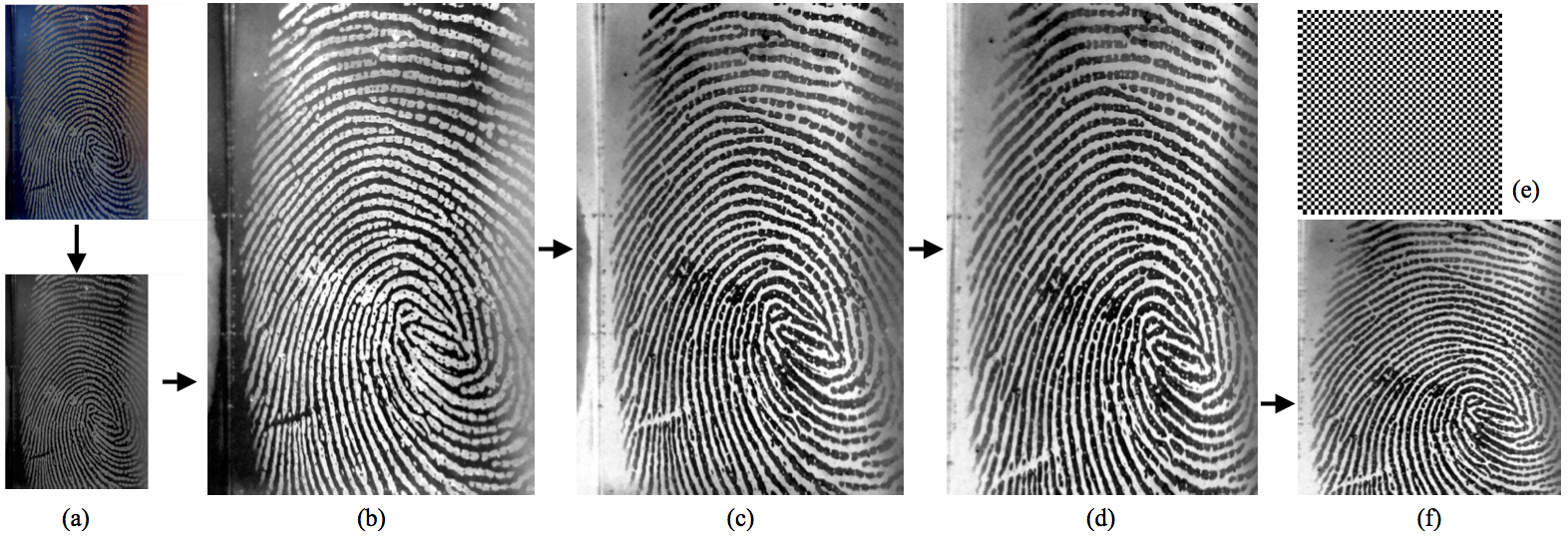}
\caption{Processing a RaspiReader raw FTIR fingerprint image into a 500 ppi fingerprint image compatible for matching with existing COTS fingerprint readers. (a) The RGB raw FTIR image is first converted to grayscale. (b) Histogram equalization is performed on the grayscale FTIR image to enhance the contrast between the fingerprint ridges and valleys. (c) The fingerprint is negated so that the ridges appear dark, and the valleys appear white. (d), (f) Calibration (estimated using the checkerboard calibration pattern in (e)) is applied to frontalize the fingerprint image to the image plane and down sample (by averaging neighborhood pixels) to 500 ppi in both the x and y directions.}
\label{fig:processing}
\vspace{-1.0em}
\end{center}
\end{figure*}

Let a raw (unprocessed) FTIR fingerprint image from the RaspiReader be denoted as $FTIR_{raw}$. This raw image $FTIR_{raw}$ is first converted from the RGB color space to grayscale ($FTIR_{gray}$) (Fig. \ref{fig:processing} (a)). Then, in order to further contrast the ridges from the valleys of the fingerprint, histogram equalization is performed on $FTIR_{gray}$ (Fig. \ref{fig:processing} (b)). Finally, $FTIR_{gray}$ is negated so that the ridges of the fingerprint image are dark, and the background of the image is white (as are fingerprint images acquired from COTS readers) (Fig. \ref{fig:processing} (c)). 


Following the aforementioned image processing techniques, the RaspiReader FTIR fingerprint images are further processed by performing a perspective transformation (to frontalize the fingerprint to the image plane) and scaling to 500 ppi (Figs. \ref{fig:processing} (d), (f)).

\begin{figure}[!h]
\begin{center}
\includegraphics[scale=0.3]{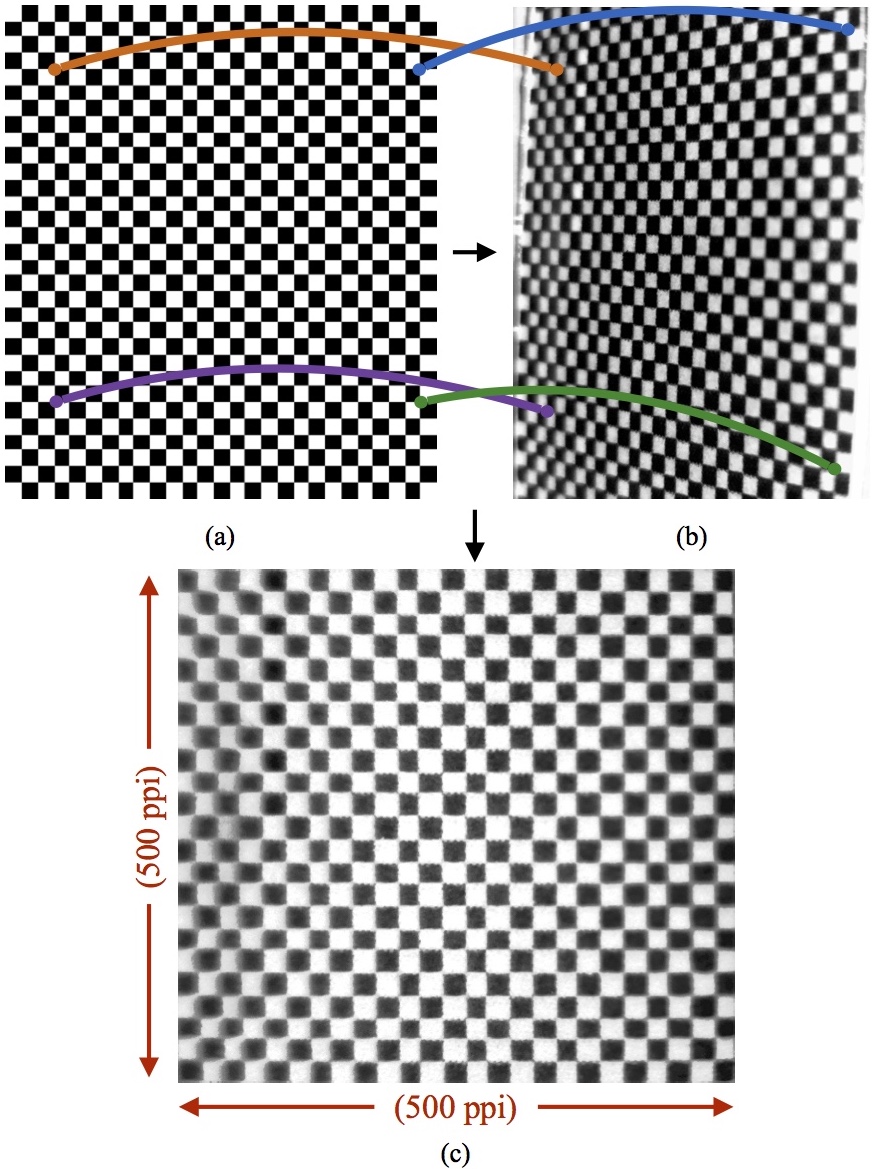}
\caption{Acquiring Image Transformation Parameters. A 2D printed checkerboard pattern (a) is imaged by the RaspiReader (b). Corresponding points between the frontalized checkerboard pattern (a) and the distorted checkerboard pattern (b) are defined so that perspective transformation parameters can be estimated to map (b) into (c). These transformation parameters are subsequently used to frontalize fingerprint images acquired by RaspiReader for the purpose of fingerprint matching. The checkerboard imaged in (b) is also used to acquire the native resolution of RaspiReader in order to scale matching images to 500 ppi in both the x and y axis as shown in (c).}
\label{fig:calibration}
\end{center}
\end{figure}

\begin{figure}[h]
  \centering
  \subfloat[]{\includegraphics[scale=.4]{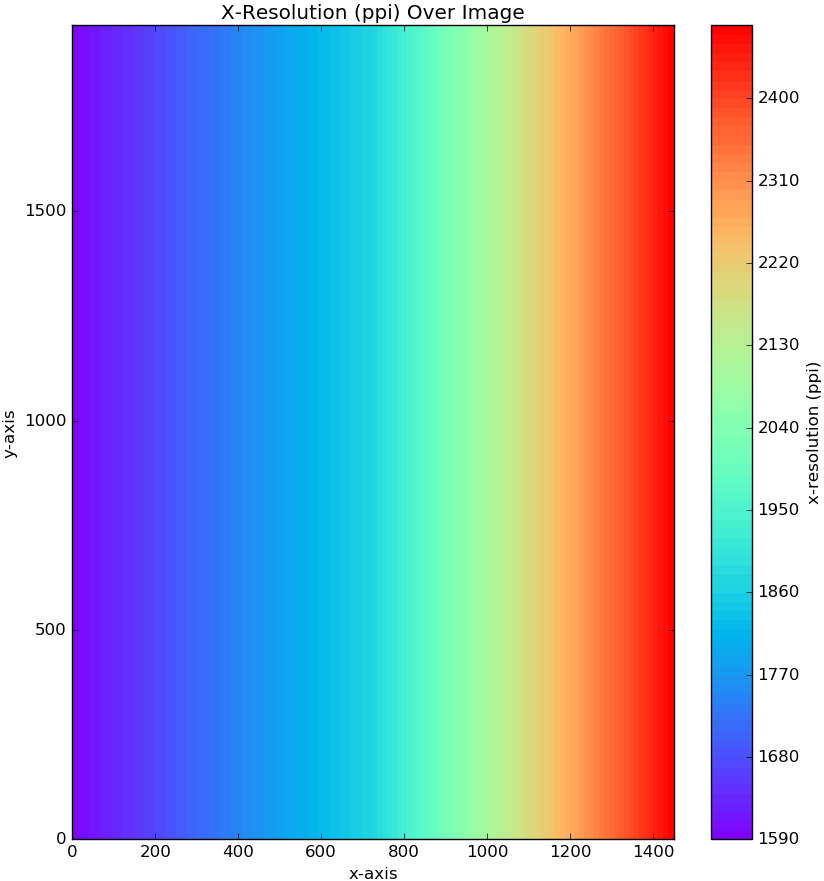}\label{fig:f1}}
  \hfill
  \subfloat[]{\includegraphics[scale=.4]{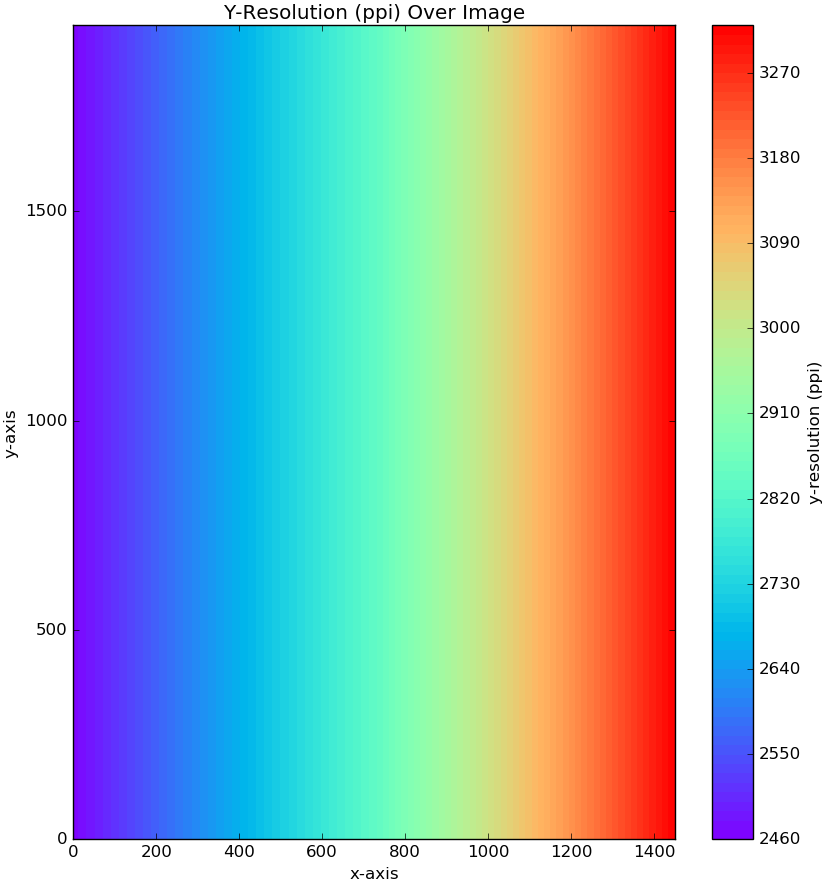}\label{fig:f2}}
  \caption{Native resolution (ppi) in (a) x-axis and (b) y-axis over the raw FTIR RaspiReader image. As is normal, native resolution changes across the image because the right side of the image is closer to the camera than the left side.}
  \label{fig:heatmap}
 \vspace{-1.0em}
\end{figure}

A perspective transformation is performed using Equation \ref{eq:2},

\begin{equation}\label{eq:2}
\begin{bmatrix}x'\\ y'\\ 1\end{bmatrix}=\frac{1}{\lambda }\begin{bmatrix} a & b & c\\ d& e & f\\ g & h & 1\end{bmatrix}\begin{bmatrix}x\\ y\\ 1\end{bmatrix}
\end{equation} where $x$ and $y$ are the source coordinates, $x'$ and $y'$ are the transformed coordinates, $(a, b, c, d, e, f, g, h)$ is the set of transformation parameters, and $\lambda = gx+hy+1$ is a scale parameter. In this work, we image a 2D printed checkerboard pattern to define source and destination coordinate pairs such that the transformation parameters could be estimated (Fig. \ref{fig:calibration}). Once the perspective transformation has been completed, the RaspiReader image is downsampled (by averaging neighborhood pixels) to 500 ppi (Fig. \ref{fig:processing} (f)). Note that the native resolution of the RaspiReader images was acquired using a 2D printed checkerboard calibration pattern (Fig. \ref{fig:calibration} (b)) and ranges from approx. 1594 ppi to 2480 ppi in the x-axis (Fig. \ref{fig:heatmap} (a)) and 2463 ppi to 3320 ppi in the y-axis (Fig. \ref{fig:heatmap} (b)). While the high resolution images captured by the RaspiReader 5 Megapixel cameras far exceed the resolution of COTS fingerprint readers (providing added minute textural details for distinguishing live fingers from spoof fingers), we observed that the focus of the native images captured by RaspiReader does deteriorate on the left and right edges (Fig. \ref{fig:calibration} (b)). We are currently investigating methods for properly focusing the lens on the entire FOV, so that minute textural details are not lost on the edges of the RaspiReader images. 

Upon completion of this entire fingerprint reader assembly and image processing procedure, the RaspiReader is fully functional and ready for use in both spoof detection and subsequent fingerprint matching.

\section{Live and Spoof Fingerprint Database Construction}

To test the utility of the RaspiReader for spoof detection and its interoperability for fingerprint matching, a database of live and spoof fingerprint impressions was collected for performing experiments. This database is constructed as follows. 

Using 7 different materials (Fig. \ref{fig:spoofs} (a)), 66 spoofs were fabricated\footnote{Our spoofs were shipped to us by Precise Biometrics \cite{precise}, a company specializing in evaluating spoof detection capability and that also has close ties to the LivDet dataset authors. As such, our spoofs are of high quality and are similar to the spoofs used in the LivDet competition.}. Then, for each of these spoofs, 10 impressions were captured at varying orientations and pressure on both the RaspiReader (Rpi) and a COTS 500 ppi optical FTIR fingerprint reader ($COTS_A$). The summary of this data collection is enumerated in Table 2.

To collect a sufficient variety of live finger data, we enlisted 15 human subjects with different skin colors (Fig. \ref{fig:spoofs} (b)). Each of these subjects gave 5 finger impressions (at different orientations and pressures) from all 10 of their fingers on both the RaspiReader and $COTS_A$ \footnote{Acquiring a fingerprint on RaspiReader involves the same user interactions that a COTS optical reader does. A user simply places their finger on a glass prism. Then, LEDs illuminate the finger surface and images are captured from both cameras over a time period of 1.5 seconds (Fig. \ref{fig:intro_fig}). The only difference in the acquisition process between a COTS reader and RaspiReader is that RaspiReader acquires two complementary images of the finger in contact with the glass platen from two separately mounted cameras.}. A summary of this data collection is enumerated in Table 3.

\begin{table}[h]
 \centering
\begin{threeparttable}
\begin{tabular}{ |c||c|c|c|c|}
 \multicolumn{5}{c}{Table 2: Summary of Spoof\tnote{1}~ Fingerprints Collected} \\
 \hline
 Material & \specialcell{Number \\of \\Spoofs\tnote{2}} &\specialcell{RPi Direct \\Images} & \specialcell{RPi FTIR \\Images} & \specialcell{$COTS_A$ \\FTIR \\Images} \\
 \hline
 \hline
 Ecoflex & 10 & 100 & 100 & 100\\
 \hline
 \specialcell{Wood Glue} & 10 & 100 & 100 & 100\\
  \hline
 \specialcell{Monster Liquid \\Latex} & 10 & 100 & 100 & 100\\
 \hline
 \specialcell{Liquid Latex \\Body Paint} & 10 & 100 & 100 & 100\\
 \hline
 Gelatin & 10 & 100 & 100 & 100\\
 \hline
 \specialcell{Silver Coated\\ Ecoflex} & 10 &  100 & 100 & 100\\
 \hline
 \specialcell{Crayola Model\\ Magic} & 6 & 60 & 60 & 60\\
 \hline
  \hline
  Total & 66 & 660 & 660 & 660\\
  \hline
\end{tabular}
\begin{tablenotes}
\item[1] The spoof materials used to fabricate these spoofs were in accordance with the approved materials by the IARPA ODIN project \cite{odin}.
\item[2] The spoofs are all of unique fingerprint patterns.
\end{tablenotes}
\end{threeparttable}
\end{table}

\begin{table}[h]
 \centering
\begin{tabular}{ |c|c|c|c|c|}
 \multicolumn{5}{c}{Table 3: Summary of Live Finger Data Collected} \\
 \hline
 \specialcell{Number of \\Subjects} & \specialcell{Number of \\Fingers} &  \specialcell{RPi Direct \\Images} &\specialcell{RPi FTIR \\Images} & \specialcell{$COTS_A$ \\FTIR\\Images} \\
 \hline
 \hline
 15 & 150 & 750 & 750 & 750\\
 \hline
\end{tabular}
\end{table}

\begin{figure}[!h]
\begin{center}
\includegraphics[scale=0.25]{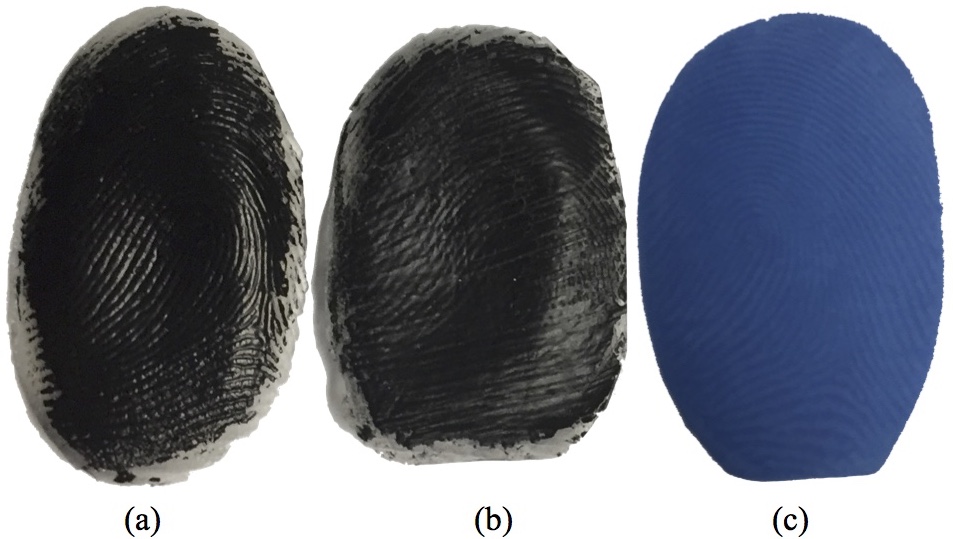}
\caption{Failure to Capture. Several spoofs are unable to be imaged by the RaspiReader due to their dissimilarity in color. In particular, because spoofs in (a) and (b) are black, all light rays will be absorbed preventing light rays from reflecting back to the FTIR imaging sensor. In (c), the dark blue color again prevents enough light from reflecting back to the camera. (a) and (b) are both ecoflex spoofs coated with two different conductive coatings. (c) is a blue crayola model magic spoof attack. }
\label{fig:failurecapture}
\vspace{-1.5em}
\end{center}
\end{figure}

In addition to the images of live finger impressions and spoof finger impressions we collected for conducting spoof detection experiments, we also verified that for spoofs with optical properties too far from that of live finger skin (Fig. \ref{fig:failurecapture}), images would not be captured by the RaspiReader. These ``failure to capture" spoofs are therefore filtered out as attacks before any software based spoof detection methods need to be performed.

\section{Spoof Detection Experiments and Results}

Given the database of live and spoof fingerprint images collected on both $COTS_A$, and the prototype RaspiReader, a number of spoof detection experiments are conducted to demonstrate the superiority of the raw images from the RaspiReader for training spoof detectors in comparison to the grayscale images output by COTS optical readers. In particular, we (i) take several successful spoof detection techniques from the literature, (ii) train and test the spoof detectors on $COTS_A$ images, (iii) train and test the spoof detectors on RaspiReader images, and (iv) compare the results to show the significant boost in performance when RaspiReader images are used to train spoof detectors rather than $COTS_A$ images. In addition, experiments are conducted to demonstrate that fingerprint images from the RaspiReader are compatible for matching with fingerprint images acquired from $COTS_A$.

\subsection{Spoof Detection Methods}

To thoroughly demonstrate the value RaspiReader images provide in training spoof detectors, we select two different spoof detection methods, namely, (i) textural features (LBP \cite{g_lbp}) in conjunction with a linear Support Vector Machine (SVM) and (ii) a Convolutional Neural Network (CNN). Textural features were chosen because of their popularity and their demonstrated superior spoof detection performance in comparison to other ``hand-crafted features" such as anatomical or physiological features in the literature \cite{2011}. CNNs were chosen as a second spoof detection method in our experiments given that they are currently state-of-the-art on the publicly available LivDet datasets. \cite{g_lbp, CNN1, CNN2, tarang}. The details of the experiments performed with both of these spoof detection methods are provided in the following subsections.

\subsubsection{LBP Features From $COTS_A$ Images} 
We begin our experiments using grayscale processed fingerprint images acquired from $COTS_A$ (Fig. \ref{fig:fingers} (c)). From these images, we extract the very prevalent grayscale and rotation invariant local binary patterns (LBP) \cite{g_lbp}. LBP features are extracted by constructing a histogram of bit string values determined by thresholding pixels in the local neighborhoods around each pixel in the image. Since image texture can be observed at different spatial resolutions, parameters $R$ and $P$ are specified in LBP construction to indicate the length (in pixels) of the neighborhood radius used for selecting pixels and also the number of neighbors to consider in a local neighborhood. Previous studies have shown that more than 90\% of fundamental textures in an image can belong to a small subset of binary patterns called ``uniform" textures (local binary patterns containing two or fewer 0/1 bit transitions) \cite{g_lbp}. Therefore, in line with previous studies using local binary patterns for fingerprint spoof detection, we also employ the use of uniform local binary patterns.

More formally, let $LBP(P,R)$ be the uniform local binary pattern histogram constructed by binning the local binary patterns for each pixel in an image according to the well known LBP operation \cite{g_lbp} with parameters $P$ and $R$. In our experiments, we extract $LBP(8,1)$, $LBP(16,2)$, and $LBP(24,3)$ in order to capture textures at different spatial resolutions. These histograms (each having $P + 2$ bins) are individually normalized and concatenated into a single feature vector $\mathbf{X}$ of dimension 54.

For classification of these features, we employ a binary linear SVM. As known in the art, an initially ``hard margin" SVM can be ``softened" by a parameter $C$ to enable better generalization of the classifier to the testing dataset. In our case, we use five-fold cross validation to select the value of $C$ (from the list of $\begin{bmatrix} 10^{-5} & 10^{-4} & ... & 10^4 & 10^5\end{bmatrix}$) such that the best performance is achieved in different folds. In our experiments, the best classification results were achieved with $C = 10^2$.

\subsubsection{CLBP Features From RaspiReader Images} 
In this experiment, we make use of the information rich images from the RaspiReader (Figs. \ref{fig:fingers} (a, b)) for spoof detection. As with Experiment 1, we again pursue the use of LBP textural features. However, since the raw images from the RaspiReader contain color information, rather than using the traditional grayscale LBP features, we employ the use of color local binary patterns (CLBP). Previous works have shown the efficacy of CLBP for both face recognition and face spoof detection \cite{color_lbp_rec, color_lbp_spoofing}. However, because fingerprint images from COTS fingerprint readers are grayscale, CLBP features have, to our knowledge, not been investigated for use in fingerprint spoof detection until now. 

\begin{figure}[!h]
\begin{center}
\includegraphics[scale=0.35]{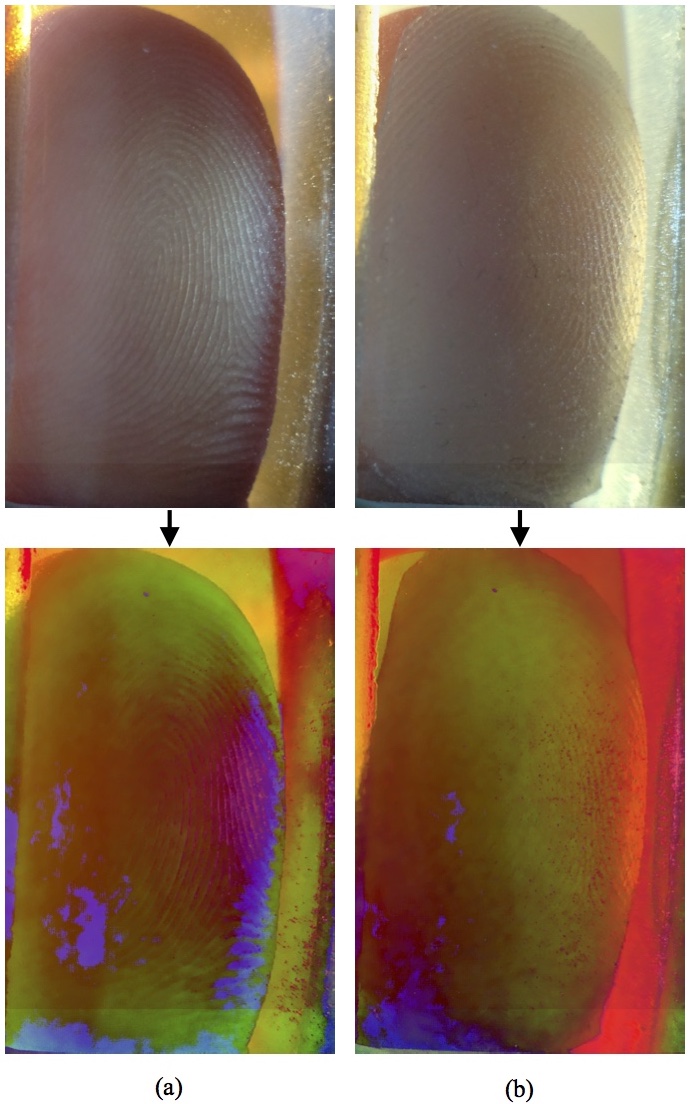}
\caption{RGB to HSV conversion. (a) A live direct view image from RaspiReader is converted to the HSV color space. (b) An ecoflex spoof attack imaged by the direct view camera of RaspiReader is converted to the HSV color space. Experimental results demonstrate a performance boost when preprocessing the RaspiReader images by converting from the RGB to HSV color space. }
\label{fig:hsv}
\vspace{-1.5em}
\end{center}
\end{figure}

Unlike traditional grayscale LBP patterns, color local binary patterns (CLBP) encode discriminative spatiochromatic textures from across multiple spectral channels \cite{color_lbp_rec}. In other words, CLBP extracts textures across all the different image bands in a given input image. More formally, given an input image $I$ with $K$ spectral channels, let the set of all spectral channels for $I$ be defined as $S = \{S_1,...,S_K\}$. Then, the CLBP feature vector $\mathbf{X}$ of dimension 486 can be extracted from $I$ using Algorithm 1. Note that in Algorithm 1, $LBP(S_i, S_j, P,R)$ returns a normalized histogram of local binary patterns using $S_i$ as the image channel that the center (thresholding) pixels are selected from, and $S_j$ as the image channel from which the neighborhood pixels are selected from in the same computation of LBP as performed in Experiment 1. Also note that in Algorithm 1, $\Vert$ indicates vector concatenation. Finally, in our experiments, we preprocess the RaspiReader input image $I$ prior to CLBP extraction by (i) downsampling (FTIR images from 1450 x 1944 to 108 x 145 and direct view images from 1290 x 1944 to 96 x 145), and (ii) converting to the HSV color space (Fig. \ref{fig:hsv})\footnote{Other color spaces were experimented with, but HSV consistently provided the highest performance. This is likely because HSV separates the luminance and chrominance components in an image, allowing extraction of features on more complementary image channels.}.

\begin{algorithm}[H] 
\caption{Extraction of Color Local Binary Patterns} 
\begin{algorithmic}
   \STATE $\mathbf{X} \gets [\text{ }]$
    \FOR {$i\gets 1, K$}
    	\FOR {$j\gets 1, K$}
		\STATE \begin{varwidth}[t]{\linewidth}
      			$\mathbf{X}$~$\gets$~$\mathbf{X} \Vert LBP(S_i, S_j, 8,1) \Vert$ \par
        			\hskip\algorithmicindent $LBP(S_i, S_j, 16,2) \Vert LBP(S_i, S_j, 24,3)$\par
      			\end{varwidth}
	\ENDFOR
    \ENDFOR
    \RETURN $\mathbf{X}$
\end{algorithmic} 
\end{algorithm} 

As in Experiment 1, a binary linear SVM with a parameter of $C = 10^2$ is trained with these features and subsequently used for classification. We again choose the parameter $C$ using 5-fold cross validation and a selection list of $\begin{bmatrix} 10^{-5} & 10^{-4} & ... & 10^4 & 10^5\end{bmatrix}$. Since RaspiReader outputs two color images (one raw FTIR image and one direct view image), we perform multiple experiments using the proposed CLBP features in conjunction with the SVM. In particular, we (i) extract CLBP features from the RaspiReader raw FTIR images to train/test a SVM, (ii) extract CLBP features from RaspiReader direct view images to train/test a SVM, and (iii) fuse CLBP features from both image outputs to train and test a SVM. We also attempted fusing CLBP features from the RaspiReader raw images with grayscale LBP features from RaspiReader processed FTIR images, but found no significant performance gains under this last fusion scheme.

\subsubsection{MobileNet}

In addition to performing experiments involving ``handcrafted" textural features, we also perform experiments where the features are directly learned and classified by a Convolutional Neural Network (CNN). In choosing a CNN architecture, we carefully considered both the size and computational overhead, since in future works, we will optimize the architecture to directly run on the RaspiReader's Raspberry Pi Processor. The need for a ``low over-head" architecture prompted us to select MobileNet \cite{mobile}. MobileNet has been shown to perform very closely (within 1 \% accuracy) to popular CNN models (VGG and Inception v3) on the ImageNet and Stanford Dogs datasets while being 32 times smaller than VGG, 7 times smaller than Inception v3, 27 times less computationally expensive than VGG, and 8 times less computationally expensive than Inception v3. 

In our experiments, we employ the Tensorflow Slim implementation of MobileNet\footnote{\url{https://github.com/tensorflow/models/tree/master/research/slim}}. MobileNet is comprised of 28 convolutional layers, and in our case, a final 2 class softmax layer for classification of live or spoof. In all of our experiments involving MobileNet, the RMSProp optimizer was used for training the network along with a batch size of 32, and an adaptive (exponential decay) learning rate. To increase the generalization ability of the networks, we employ various data augmentation methods such as brightness adjustment, random cropping, and horizontal and vertical reflections.

Using the aforementioned MobileNet architecture and hyper-parameters, we train/test the network with (i) $COTS_A$ grayscale fingerprint images, (ii) RaspiReader raw FTIR images, (iii) RaspiReader direct view images, and (iv) RaspiReader processed FTIR images. Additionally, we perform experiments in which we fuse the score outputs of MobileNet models trained on the different image outputs from RaspiReader to take advantage of the complementary information within the different RaspiReader image outputs. When training and testing MobileNet with $COTS_A$ images or RaspiReader processed FTIR images, the three input channels of the network are each fed with the same down sampled (357 x 392 to 224 x 224) grayscale $COTS_A$ image or (290 x 267 to 224 x 224) RaspiReader processed FTIR image. When training the network with RaspiReader raw images, we again down sample the images (1450 x 1944 to 224 x 224 for raw FTIR and 1290 x 1944 to 224 x 224 for direct image), however, in this case, each of the three color channels are fed as input to the three input channels of the network. More specifically, we first convert the RaspiReader image to HSV (given our earlier findings of superior performance in this color space), and then feed each channel H, S, and V into the network's input channels.

\subsection{Spoof Detection Results}

Using the spoof detection schemas previously described, we train and test classifiers under two main scenarios. In the first scenario, we train the classifier on a subset of spoof images from every type in the dataset (Table 2). During testing, spoof images from the same spoof types seen during training will be passed to the spoof detector for classification.  We hereafter refer to this training and testing scenario as a ``known-material" scenario. In the second scenario, we train the classifier with images from all of the spoof types in the dataset except one (i.e. the spoof impressions from one type of spoof are withheld). Then, during testing the impressions of the withheld spoof type are used for testing. In the literature, this type of spoof detection evaluation is referred to as a ``cross-material" scenario. In the following experimental results, we demonstrate that the RaspiReader images significantly boost the spoof detection performance in both the known-material evaluations and the cross-material evaluations. 

\subsubsection{Known-Material Scenarios} The first known-material results are reported in accordance with spoof detection methods 1 and 2. That is, we extract textural features from both $COTS_A$ images and RaspiReader images respectively, train and test linear SVMs, and finally, compare the results (Table 4). In all of our known-material scenario experiments, we report the average spoof detection performance and standard deviation over 5-folds. That is, for spoof data, we select 80\% of the spoof impressions from each spoof material for training (each fold) and use the remaining 20\% for testing.  For live finger data, we select the finger impressions of 12 subjects each fold (600 total images) for training, and use the live finger impressions of the remaining 3 subjects for testing.

\begin{table}[h]
 \centering
 \begin{threeparttable}
\begin{tabular}{ |c||c|c|}
 \multicolumn{3}{c}{Table 4: Textural Features and Known Testing Materials} \\
 \hline
 Method & \specialcell{TDR @ FDR = 1.0\% \\ $\mu \pm \sigma$\tnote{1}} & \specialcell{Detection Time (msecs)} \\
 \hline
 \hline
 \specialcell{$COTS_A$ \\+ LBP} & $75.9\% \pm 30.8$ & 236 \\
 \hline
 \specialcell{Rpi raw FTIR \\+ CLBP} & $91.5\% \pm 11.0$ & 243 \\
  \hline
 \specialcell{Rpi Direct \\+ CLBP} & $98.10\% \pm 1.9$ & 243 \\
 \hline
 \specialcell{Rpi Fusion \\+ CLBP\tnote{2}} & $97.7\% \pm 3.0$ & 486 \\
 \hline
\end{tabular}
\begin{tablenotes}
\item[1] These results are reported over 5-folds.
\item[2] Rpi Fusion + CLBP is a feature level fusion (concatenation) of CLBP features extracted from both Rpi raw FTIR images and Rpi Direct Images, respectively.
\end{tablenotes}
\end{threeparttable}
\end{table}

From the results of Table 4, one can observe that both image outputs of the RaspiReader contain far more discriminative information for spoof detection than the processed grayscale images output by $COTS_A$. In particular, spoof detection performance is significantly higher when extracting textural (CLBP) features from the RaspiReader images, than when extracting textural features (LBP) from $COTS_A$ images. While in these first results, the fusion of features from both RaspiReader image outputs actually hurts the classification performance slightly (compared to extracting features only from the direct view images), in subsequent experiments, we will demonstrate that different feature extraction and classification techniques can better utilize the multiple outputs of RaspiReader in a complementary manner to instead boost the classification performance.

The second known-material results are reported in accordance with spoof detection scheme 3. More specifically, the results are reported (over 5-folds) when MobileNet is trained and tested with (i) $COTS_A$ images, (ii) RaspiReader processed FTIR images, (iii) RaspiReader raw FTIR images, and (iv) RaspiReader direct images. In addition, we report the results when fusing the score outputs of multiple MobileNet models trained on the different image outputs of RaspiReader (Table 5). 

\begin{table}[h]
 \centering
 \begin{threeparttable}
\begin{tabular}{ |c||c|c|}
 \multicolumn{3}{c}{Table 5: MobileNet and Known Testing Materials} \\
 \hline
 Method & \specialcell{TDR @ FDR = 1.0\% \\ $\mu \pm \sigma$\tnote{1}} & \specialcell{Detection Time (msecs)} \\
 \hline
 \hline
 \specialcell{$COTS_A$ \\+ MobileNet} & $91.9\% \pm 8.0$ & 22 \\
  \hline
 \specialcell{Rpi processed FTIR\\+ MobileNet} & $94.5\% \pm 3.7$ & 22 \\
 \hline
 \specialcell{Rpi raw FTIR \\+ MobileNet} & $95.1\% \pm 5.6$ & 22 \\
  \hline
 \specialcell{Rpi Direct \\+ MobileNet} & $95.3\% \pm 3.5$ & 22 \\
 \hline
 \specialcell{Rpi Fusion 2 \\+ MobileNet\tnote{2}} & $98.4\% \pm 2.3$ & 45 \\
  \hline
 \specialcell{Rpi Fusion 3 \\+ MobileNet\tnote{3}} & $98.9\% \pm 1.5$ & 67 \\
 \hline
\end{tabular}
\begin{tablenotes}
\item[1] These results are reported over 5-folds.
\item[2] Rpi Fusion 2 + MobileNet is a score level fusion (averaging) of a MobileNet model trained on Rpi raw FTIR images and a MobileNet model trained on Rpi Direct Images.
\item[3] Rpi Fusion 3 + MobileNet is a score level fusion (averaging) of separate MobileNet models trained on Rpi raw FTIR images, Rpi Direct Images, and on Rpi processed FTIR images.
\end{tablenotes}
\end{threeparttable}
\end{table}

The results of Table 5 show that both the raw image outputs of RaspiReader and the processed image output of RaspiReader contain more discriminative information for spoof detection than the processed images output by $COTS_A$. The MobileNet models trained on RaspiReader images always outperform the MobileNet model trained on $COTS_A$ grayscale images both in average spoof detection performance and stability (significantly lower s.d.). What is further interesting about the results of Table 5 is that the features extracted by MobileNet from each RaspiReader output are quite complementary, demonstrated by the fact that spoof detection performance is improved when fusing the scores of MobileNet models trained on each RaspiReader image output. So, while CLBP features outperform MobileNet on the RaspiReader direct images, the fused MobileNet classifiers outperform the fused CLBP classifier. 

\subsubsection{Cross-Material Scenarios}

The cross-material results use the same spoof detection schemas as enumerated in the known-material results with a primary difference being the training and testing data splits provided to the various classifiers. In all the cross-material scenarios, spoof impressions of six materials are partitioned to the classifier for training, and the spoof impressions of one ``unseen" material are kept aside for testing. In this manner the generalization capability of the spoof detector to novel spoof types is thoroughly assessed. For live finger data, we randomly select the finger impressions of two subjects (100 total images) for testing, and use the live finger impressions of the remaining thirteen subjects for training. Since there are seven different spoof materials in our training set (Table 2), we conduct seven different cross-material experiments for each spoof detection schema (where one of the seven spoof types is left aside for testing). The cross material results when using textural features in conjunction with SVMs is reported in Table 6. The cross-material results when using MobileNet extracted features is reported in Table 7. Note, we only report the best textural fusion and CNN fusion methods in the cross-material results. The other non-fusion based methods were experimented with, but did not provide as high of performance in the cross-material scenarios.

\begin{table}[h]
 \centering
 \begin{threeparttable}
\begin{tabular}{ |c||c|c|}
 \multicolumn{3}{c}{Table 6: Textural Features and Cross-Material Testing\tnote{1}} \\
 \hline
 Testing Material & \specialcell{$COTS_A$ \\+ LBP} & \specialcell{Rpi Fusion + CLBP\tnote{2}} \\
 \hline
 \hline
 \specialcell{Crayola Model \\ Magic} & $91.7\%$ & $98.3\%$ \\
 \hline
 \specialcell{Ecoflex} & $66.0\%$ & $77.0\%$ \\
  \hline
 \specialcell{Silver Coated \\Ecoflex} & $88.0\%$ & $100.0\%$ \\
 \hline
 \specialcell{Gelatin} & $62.0\%$ & $87.0\%$ \\
 \hline
  \specialcell{Liquid Latex \\ Body Paint} & $84.0\%$ & $100.0\%$ \\
 \hline
  \specialcell{Monster Liquid \\Latex} & $68.0\%$ & $98.0\%$ \\
 \hline
  \specialcell{Wood Glue} & $100.0\%$ & $81.0\%$ \\
 \hline
\end{tabular}
\begin{tablenotes}
\item[1] TDR @ FDR = 1.0\% is reported
\item[2] Rpi Fusion + CLBP is a feature level fusion (concatentation) of CLBP features extracted from both Rpi raw FTIR images and Rpi Direct Images, respectively.
\end{tablenotes}
\end{threeparttable}
\end{table}

\begin{table}[h]
 \centering
 \begin{threeparttable}
\begin{tabular}{ |c||c|c|c|}
 \multicolumn{4}{c}{Table 7: MobileNet and Cross-Material Testing\tnote{1}} \\
 \hline
 Testing Material & \specialcell{$COTS_A$ \\+ MobileNet} & \specialcell{Rpi \\Fusion 2 + \\MobileNet\tnote{2}} &  \specialcell{Rpi \\Fusion 3 + \\MobileNet\tnote{3}}\\
 \hline
 \hline
 \specialcell{Crayola Model \\ Magic} & $50.0\%$ & $100.0\%$ & $100.0\%$\\
 \hline
 \specialcell{Ecoflex} & $100.0\%$ & $8.0\%$ & $56.0\%$ \\
  \hline
 \specialcell{Silver Coated \\Ecoflex} & $77.0\%$ & $100.0\%$ & $100.0\%$\\
 \hline
 \specialcell{Gelatin} & $88.0\%$ & $100.0\%$ & $100.0\%$\\
 \hline
  \specialcell{Liquid Latex \\ Body Paint} & $97.0\%$ & $100.0\%$ & $100.0\%$\\
 \hline
  \specialcell{Monster Liquid \\Latex} & $86.0\%$ & $100.0\%$ & $100.0\%$\\
 \hline
  \specialcell{Wood Glue} & $94.0\%$ & $96.0\%$ & $96.0\%$\\
 \hline
\end{tabular}
\begin{tablenotes}
\item[1] TDR @ FDR = 1.0\% is reported
\item[2] Rpi Fusion 2 + MobileNet is a score level fusion (max) of separate MobileNet models trained on Rpi raw FTIR images and on Rpi Direct Images, respectively.
\item[3] Rpi Fusion 3 + MobileNet is a score level fusion (max) of separate MobileNet models trained on Rpi raw FTIR images, Rpi Direct Images, and on Rpi processed FTIR images.
\end{tablenotes}
\end{threeparttable}
\end{table}

The key findings of the cross-material experiments as revealed in Tables 6 and 7 are as follows. First, in both textural based spoof detection methods and CNN based spoof detection methods, the raw images output by RaspiReader provide more discriminative information than COTS grayscale fingerprint images. This enables much higher spoof detection performance on spoofs fabricated from materials not seen by the classifier during training, a major flaw in many existing spoof detection methods relying on only COTS grayscale images.

The one case of poor cross-material performance (when using RaspiReader images) came when the testing material withheld was ecoflex (Table 7). This can be explained by ecoflex being a very transparent spoof, enabling much of the live finger color behind the spoof to seep through. As such, when the MobileNet models were trained on the other non-transparent spoofs and tested on the transparent ecoflex, the performance dropped considerably. However, we also noticed that the best cross-material performance (when using $COTS_A$ images) came when the testing material withheld was ecoflex. The most plausible explanation for this is that the MobileNet model trained on the $COTS_A$ images must focus on textural features rather than color. As such, the transparent property of ecoflex did not affect the classifier trained on the grayscale images. This prompted us to train a third model on the RaspiReader processed FTIR images (i.e. the raw FTIR images were converted to grayscale and contrast enhanced). We then fused the score of this third model with the two MobileNet models trained on the RaspiReader raw FTIR and direct images respectively. The final product was a three CNN model system which performed much better on the ecoflex testing scenario (48\% improvement). While the ecoflex testing scenario is still low, in a real world setting, this limitation is easily solved by including one transparent spoof in the training set (evidenced by the fact that in the known-material experiments, ecoflex could be differentiated from live fingers with high accuracy).

\section{Interoperability of RaspiReader}

In addition to demonstrating the usefulness of the RaspiReader images for fingerprint spoof detection, we also demonstrate that by processing the RaspiReader FTIR images, we can output images which are compatible for matching with images from COTS fingerprint readers. Previously, we discussed how to process and transform a RaspiReader raw FTIR image into an image suitable for matching. In this experiment, we evaluate the matching performance (of 11,175 imposter pairs and 6,750 genuine pairs) when using (i) the RaspiReader processed images as both the enrollment and probe images, (ii) the $COTS_A$ images as both the enrollment and probe images, and (iii) the $COTS_A$ images as the enrollment images and the RaspiReader processed images as the probe images. The results for these matching experiments are listed in Table 8. 

\begin{table}[h]
 \centering
  \begin{threeparttable}
\begin{tabular}{ |c|c|c|}
 \multicolumn{3}{c}{\specialcell{Table 8: Fingerprint Matching Results\tnote{1}}} \\
 \hline
 Enrollment Reader & \specialcell{Probe Reader} &\specialcell{TAR @ FAR = 0.1\%} \\
 \hline
 \hline
 \specialcell{$COTS_A$} & $COTS_A$ & 98.62\% \\
 \hline
 \specialcell{RaspiReader} & RaspiReader & 99.21\% \\
  \hline
 \specialcell{$COTS_A$} & RaspiReader & 95.56\% \\
 \hline
\end{tabular}
\begin{tablenotes}
\item[1] We use the Innovatrics fingerprint SDK which is shown to have high accuracy in the NIST FpVTE evaluation \cite{nist}.
\end{tablenotes}
\end{threeparttable}
\end{table}

From these results, we make two observations. First, the best performance is achieved for native comparisons, where the enrolled and search (probe) images are produced by the same capture device. RaspiReader's native performance is slightly better than that of $COTS_A$. This indicates that the RaspiReader is capable of outputting images which are compatible with state of the art fingerprint matchers. Second, we note that the performance does drop slightly when conducting the interoperability experiment ($COTS_A$ is used for enrollment images and RaspiReader is used for probe images). However, the matching performance is still quite high considering the stringent operating point (FAR = 0.1\%). Furthermore, studies have shown that when different fingerprint readers are used for enrollment and subsequent verification or identification, the matching performance indeed drops \cite{ross_cross,cross,engelsma2017universal}. Finally, we are currently investigating other approaches for processing and downsampling RaspiReader images to reduce some of the drop in cross-reader performance.

\section{Computational Resources} 

All image preprocessing, LBP and CLBP feature extractions, and SVM classifications were performed with a single CPU core on a Macbook Pro running a 2.9 GHz Intel Core i5 processor. MobileNet training and classification was performed on a single Nvidia GTX Titan GPU. The total time from image capture to spoof detection with our best MobileNet model (RpiFusion3) is approximately 3.067 seconds (1.5 seconds for image capture, 1.5 seconds to transmit data to GPU, and 67 milliseconds for classification). In the future, we will port all spoof detection and fingerprint matching onto the RaspiReader creating a completely portable and secure ``fingerprint match on box".

\section{Conclusions}
We have open sourced\footnote{\url{https://github.com/engelsjo/RaspiReader}}, the design and assembly of a custom fingerprint reader, called RaspiReader, with Raspberry Pi and other ubiquitous components. This fingerprint reader is both low cost (US \$175) and easy to assemble, enabling other researchers to easily and seamlessly develop their own novel fingerprint spoof detection solutions which use both hardware and software. By customizing RaspiReader with two cameras for fingerprint image acquisition rather than the customary one, we were able to extract discriminative information from both raw images which, when fused together, enabled us to achieve higher spoof detection performance (in both known-material and cross-material testing scenarios) compared to when features were extracted from COTS grayscale images. Finally, by processing the raw FTIR images of the RaspiReader, we were able to output fingerprint images compatible for matching with COTS optical fingerprint readers demonstrating the interoperability of RaspiReader. 

In our ongoing work, we plan to integrate specialized hardware into RaspiReader such as Optical Coherence Tomography (OCT) for sub-dermal imagery, IR cameras for vein detection, or microscopes for capturing extremely high resolution images of the fingerprint. Because the RaspiReader uses ubiquitous components running open source software, RaspiReader enables future integration of these additional hardware components. In addition to the integration of specialized hardware, we also plan to use the raw, information rich images from the RaspiReader to pursue one-class classification schemes for fingerprint spoof detection. In particular, we posit that the RaspiReader images will assist us in modeling the class of live fingerprint images, such that spoofs of all material types can be easily rejected. Finally, we will make RaspiReader a self contained fingerprint recognition system (similar to ``match on card \cite{match_on_card}"), so that fingerprint image acquisition, spoof detection, feature extraction, and matching can all be accomplished inside RaspiReader. This will provide an entire, portable, secure ``fingerprint match in a box" on an approximately 4 inch cube.


%

\ifCLASSOPTIONcompsoc
\section*{Acknowledgment}
This research was supported by grant no. 70NANB17H027 from the NIST Measurement Science program and by the Office of the Director of National Intelligence (ODNI), Intelligence Advanced Research Projects Activity (IARPA), via IARPA R\&D Contract No. 2017 - 17020200004. The
views and conclusions contained herein are those of the
authors and should not be interpreted as necessarily representing
the official policies, either expressed or implied,
of ODNI, IARPA, or the U.S. Government. The U.S. Government
is authorized to reproduce and distribute reprints
for governmental purposes notwithstanding any copyright
annotation therein.
\else
\section*{Acknowledgment}
This research was supported by both grant no. 70NANB17H027 from the NIST Measurement Science program and also by the Office of the Director of National Intelligence (ODNI), Intelligence Advanced Research Projects Activity (IARPA), via IARPA R\&D Contract No. 2017 - 17020200004. The
views and conclusions contained herein are those of the
authors and should not be interpreted as necessarily representing
the official policies, either expressed or implied,
of ODNI, IARPA, or the U.S. Government. The U.S. Government
is authorized to reproduce and distribute reprints
for governmental purposes notwithstanding any copyright
annotation therein.
\fi

\ifCLASSOPTIONcaptionsoff
  \newpage
\fi



%
\bibliography{cites}

\begin{thebibliography}{10}

\bibitem{iso}
``\uppercase{I}nternational \uppercase{S}tandards \uppercase{O}rganization,
  “\uppercase{ISO/IEC} 30107-1:2016, \uppercase{I}nformation
  \uppercase{T}echnology \uppercase{B}iometric \uppercase{P}resentation
  \uppercase{A}ttack \uppercase{D}etection \uppercase{P}art 1:
  \uppercase{F}ramework”.'' \url{https://www.iso.org/standard/53227.html,},
  2016.

\bibitem{odin}
``\uppercase{IARPA} \uppercase{Odin} program.''
  \url{https://www.iarpa.gov/index.php/research-programs/odin/odin-baa}.

\bibitem{spoofs_survey}
E.~Marasco and A.~Ross, ``A survey on antispoofing schemes for fingerprint
  recognition systems,'' {\em ACM Comput. Surv.}, vol.~47, pp.~28:1--28:36,
  Nov. 2014.

\bibitem{handbook}
D.~Maltoni, D.~Maio, A.~K. Jain, and S.~Prabhakar, {\em Handbook of Fingerprint
  Recognition}.
\newblock Springer, 2nd~ed., 2009.

\bibitem{altered_yoon}
S.~Yoon, J.~Feng, and A.~K. Jain, ``Altered fingerprints: Analysis and
  detection,'' {\em IEEE Transactions on Pattern Analysis and Machine
  Intelligence}, vol.~34, no.~3, pp.~451--464, 2012.

\bibitem{gummy}
T.~Matsumoto, H.~Matsumoto, K.~Yamada, and S.~Hoshino, ``Impact of artificial
  gummy fingers on fingerprint systems,'' in {\em Proceedings of SPIE},
  vol.~4677, pp.~275--289, 2002.

\bibitem{india1}
``\uppercase{U}nique \uppercase{I}dentification \uppercase{A}uthority of
  \uppercase{I}ndia, dashboard summary.''
  \url{https://portal.uidai.gov.in/uidwebportal/dashboard.do}.

\bibitem{india2}
``\uppercase{UIDAI} biometric device specifications (authentication).''
  \url{http://www.stqc.gov.in/sites/upload_files/stqc/files/UIDAI-Biometric-Device-Specifications-Authentication-14-05-2012_0.pdf}.

\bibitem{india3}
``\uppercase{UPI} \uppercase{U}nited \uppercase{P}ayments
  \uppercase{I}nterface.''
  \url{http://www.npci.org.in/documents/UPI_Procedural_Guidelines.pdf}.

\bibitem{obim}
``\uppercase{O}ffice of \uppercase{B}iometric \uppercase{I}dentity
  \uppercase{M}anagement.'' \url{https://www.dhs.gov/obim}.

\bibitem{hongkongchina}
``\uppercase{R}eport \uppercase{H}ong \uppercase{K}ong \uppercase{C}hina
  \uppercase{B}order \uppercase{B}iometrics \uppercase{D}evice
  \uppercase{S}poofed.''
  \url{http://cw.com.hk/news/report-hong-kong-china-border-biometrics-device-spoofed}.

\bibitem{caophone}
K.~Cao and A.~K. Jain, ``Hacking mobile phones using 2d printed fingerprints,''
  tech. rep., MSU Technical report, MSU-CSE-16-2, 2016.

\bibitem{com1}
K.~Nixon, V.~Aimale, and R.~Rowe, ``Spoof detection schemes,'' {\em Handbook of
  Biometrics}, pp.~403--423, 2008.

\bibitem{com3}
``Goodix live finger detection.''
  \url{https://findbiometrics.com/goodix-zte-biometric-sensor-3103187/}.

\bibitem{blood_flow}
P.~D. Lapsley, J.~A. Lee, D.~F. Pare~Jr, and N.~Hoffman, ``Anti-fraud biometric
  scanner that accurately detects blood flow,'' Apr.~7 1998.
\newblock US Patent 5,737,439.

\bibitem{odor}
D.~Baldisserra, A.~Franco, D.~Maio, and D.~Maltoni, ``Fake fingerprint
  detection by odor analysis,'' in {\em International Conference on
  Biometrics}, pp.~265--272, Springer, 2006.

\bibitem{rowe1}
R.~K. Rowe, ``Multispectral imaging biometrics,'' Dec.~2 2008.
\newblock US Patent 7,460,696.

\bibitem{oct}
A.~Shiratsuki, E.~Sano, M.~Shikai, T.~Nakashima, T.~Takashima, M.~Ohmi, and
  M.~Haruna, ``Novel optical fingerprint sensor utilizing optical
  characteristics of skin tissue under fingerprints,'' in {\em Biomedical
  Optics 2005}, pp.~80--87, International Society for Optics and Photonics,
  2005.

\bibitem{hardware_issues}
M.~Sepasian, C.~Mares, and W.~Balachandran, ``Liveness and spoofing in
  fingerprint identification: Issues and challenges,'' in {\em Proceedings of
  the 4th WSEAS International Conference on Computer Engineering and
  Applications}, CEA'10, pp.~150--158, 2009.

\bibitem{burned}
T.~Van~der Putte and J.~Keuning, ``Biometrical fingerprint recognition: don’t
  get your fingers burned,'' in {\em Smart Card Research and Advanced
  Applications}, pp.~289--303, Springer, 2000.

\bibitem{schuckers}
S.~A. Schuckers, ``Spoofing and anti-spoofing measures,'' {\em Information
  Security technical report}, vol.~7, no.~4, pp.~56--62, 2002.

\bibitem{texture0}
S.~B. Nikam and S.~Agarwal, ``Local binary pattern and wavelet based spoof
  fingerprint detection,'' {\em Int. J. Biometrics}, vol.~1, pp.~141--159, Aug.
  2008.

\bibitem{texture1}
L.~Ghiani, G.~L. Marcialis, and F.~Roli, ``Fingerprint liveness detection by
  local phase quantization,'' in {\em Pattern Recognition (ICPR), 2012 21st
  International Conference on}, pp.~537--540, IEEE, 2012.

\bibitem{texture2}
L.~Ghiani, A.~Hadid, G.~L. Marcialis, and F.~Roli, ``Fingerprint liveness
  detection using binarized statistical image features,'' in {\em Biometrics:
  Theory, Applications and Systems (BTAS), 2013 IEEE Sixth International
  Conference on}, pp.~1--6, IEEE, 2013.

\bibitem{texture3}
D.~Gragnaniello, G.~Poggi, C.~Sansone, and L.~Verdoliva, ``Fingerprint liveness
  detection based on weber local image descriptor,'' in {\em Biometric
  Measurements and Systems for Security and Medical Applications (BIOMS), 2013
  IEEE Workshop on}, pp.~46--50, IEEE, 2013.

\bibitem{texture4}
D.~Gragnaniello, G.~Poggi, C.~Sansone, and L.~Verdoliva, ``Local contrast phase
  descriptor for fingerprint liveness detection,'' {\em Pattern Recognition},
  vol.~48, no.~4, pp.~1050--1058, 2015.

\bibitem{pores}
L.~Ghiani, P.~Denti, and G.~L. Marcialis, ``Experimental results on fingerprint
  liveness detection,'' in {\em Proceedings of the 7th International Conference
  on Articulated Motion and Deformable Objects}, AMDO'12, pp.~210--218,
  Springer-Verlag, 2012.

\bibitem{perspiration1}
A.~Abhyankar and S.~Schuckers, ``Integrating a wavelet based perspiration
  liveness check with fingerprint recognition,'' {\em Pattern Recogn.},
  vol.~42, pp.~452--464, Mar. 2009.

\bibitem{perspiration2}
E.~Marasco and C.~Sansone, ``Combining perspiration-and morphology-based static
  features for fingerprint liveness detection,'' {\em Pattern Recognition
  Letters}, vol.~33, no.~9, pp.~1148--1156, 2012.

\bibitem{CNN1}
D.~Menotti, G.~Chiachia, A.~da~Silva~Pinto, W.~R. Schwartz, H.~Pedrini, A.~X.
  Falc{\~{a}}o, and A.~Rocha, ``Deep representations for iris, face, and
  fingerprint spoofing detection,'' {\em {IEEE} Trans. Information Forensics
  and Security}, vol.~10, no.~4, pp.~864--879, 2015.

\bibitem{CNN2}
R.~Nogueira, R.~Lotufo, and R.~Machado, ``Fingerprint liveness detection using
  convolutional neural networks,'' {\em {IEEE} Trans. Information Forensics and
  Security}, vol.~11, no.~6, pp.~1206--1213, 2016.

\bibitem{tarang}
T.~Chugh, K.~Cao, and A.~Jain, ``Fingerprint spoof detection using
  minutiae-based local patches,'' in {\em 2017 IEEE International Joint
  Conference on Biometrics (IJCB)}, IEEE, 2017.

\bibitem{2015}
V.~Mura, L.~Ghiani, G.~L. Marcialis, F.~Roli, D.~A. Yambay, and S.~A.~C.
  Schuckers, ``Livdet 2015 fingerprint liveness detection competition 2015.,''
  in {\em BTAS}, pp.~1--6, IEEE, 2015.

\bibitem{inter1}
E.~Marasco and C.~Sansone, ``On the robustness of fingerprint liveness
  detection algorithms against new materials used for spoofing,'' in {\em Proc.
  Int. Conf. Bio-Inspired Syst. Signal Process.}, pp.~553--558, 2011.

\bibitem{inter2}
B.~Tan, A.~Lewicke, D.~Yambay, and S.~Schuckers, ``The effect of environmental
  conditions and novel spoofing methods on fingerprint anti-spoofing
  algorithms,'' {\em {IEEE} Information Forensics and Security, {WIFS}}, 2010.

\bibitem{2011}
D.~Yambay, L.~Ghiani, P.~Denti, G.~L. Marcialis, F.~Roli, and S.~A.~C.
  Schuckers, ``Livdet 2011 - fingerprint liveness detection competition
  2011.,'' in {\em Proc. International Conf. Biometrics}, pp.~208--215, IEEE,
  2012.

\bibitem{open1}
A.~Rattani, W.~J. Scheirer, and A.~Ross, ``Open set fingerprint spoof detection
  across novel fabrication materials,'' {\em IEEE Trans. on Information
  Forensics and Security}, vol.~10, no.~11, pp.~2447--2460, 2015.

\bibitem{open2}
A.~Rattani and A.~Ross, ``Automatic adaptation of fingerprint liveness detector
  to new spoof materials,'' in {\em 2014 IEEE International Joint Conference on
  Biometrics (IJCB)}, pp.~1--8, IEEE, 2014.

\bibitem{open3}
Y.~Ding and A.~Ross, ``An ensemble of one-class svms for fingerprint spoof
  detection across different fabrication materials,'' in {\em {IEEE}
  International Workshop on Information Forensics and Security, {WIFS} 2016,
  Abu Dhabi, December 4-7, 2016}, pp.~1--6, 2016.

\bibitem{nist}
C.~I. Watson, G.~Fiumara, E.~Tabassi, S.~Cheng, P.~Flanagan, and W.~Salamon,
  ``Fingerprint vendor technology evaluation, nist interagency/internal report
  8034: 2015,''
\newblock available at \url{https://dx.doi.org/10.6028/NIST.IR.8034}.

\bibitem{thor}
``\uppercase{T}hor\uppercase{L}abs.''
  \url{https://www.thorlabs.com/thorproduct.cfm?partnumber=PS911}.

\bibitem{meshlab}
P.~Cignoni, M.~Corsini, and G.~Ranzuglia, ``Meshlab: an open-source 3d mesh
  processing system,'' {\em ERCIM News}, pp.~45--46, April 2008.

\bibitem{multi}
``Multi camera adapter module for raspberry pi.''
  \url{https://www.arducam.com/multi-camera-adapter-module-raspberry-pi}.
\newblock Accessed: 2017-4-15.

\bibitem{precise}
``\uppercase{P}recise \uppercase{B}iometrics.''
  \url{https://precisebiometrics.com/products/fingerprint-spoof-liveness-detection/sensor-vulnerability-analysis/}.

\bibitem{g_lbp}
T.~Ojala, M.~Pietik\"{a}inen, and T.~M\"{a}enp\"{a}\"{a}, ``Multiresolution
  gray-scale and rotation invariant texture classification with local binary
  patterns,'' {\em IEEE Trans. Pattern Anal. Mach. Intell.}, vol.~24,
  pp.~971--987, July 2002.

\bibitem{color_lbp_rec}
J.~Y. Choi, K.~N. Plataniotis, and Y.~M. Ro, ``Using colour local binary
  pattern features for face recognition,'' in {\em Proceedings of the
  International Conference on Image Processing, {ICIP} 2010, September 26-29,
  Hong Kong}, pp.~4541--4544, 2010.

\bibitem{color_lbp_spoofing}
Z.~Boulkenafet, J.~Komulainen, and A.~Hadid, ``Face spoofing detection using
  colour texture analysis,'' {\em {IEEE} Trans. Information Forensics and
  Security}, vol.~11, no.~8, pp.~1818--1830, 2016.

\bibitem{mobile}
A.~G. Howard, M.~Zhu, B.~Chen, D.~Kalenichenko, W.~Wang, T.~Weyand,
  M.~Andreetto, and H.~Adam, ``Mobilenets: Efficient convolutional neural
  networks for mobile vision applications,'' {\em arXiv preprint
  arXiv:1704.04861}, 2017.

\bibitem{ross_cross}
A.~Ross and A.~Jain, ``Biometric sensor interoperability: A case study in
  fingerprints,'' in {\em ECCV Workshop BioAW}, pp.~134--145, Springer, 2004.

\bibitem{cross}
X.~Jia, X.~Yang, Y.~Zang, N.~Zhang, and J.~Tian, ``A cross-device matching
  fingerprint database from multi-type sensors,'' in {\em Pattern Recognition
  (ICPR), 2012 21st International Conference on}, pp.~3001--3004, IEEE, 2012.

\bibitem{engelsma2017universal}
J.~J. Engelsma, S.~S. Arora, A.~K. Jain, and N.~G. Paulter~Jr, ``Universal 3d
  wearable fingerprint targets: Advancing fingerprint reader evaluations,''
  {\em arXiv preprint arXiv:1705.07972}, 2017.

\bibitem{match_on_card}
``\uppercase{S}ecure\uppercase{ID} \uppercase{N}ews: match-on-card
  biometrics.''
  \url{https://www.secureidnews.com/news-item/tech-101-match-on-card-biometrics/}.

\end{thebibliography}
\bibliographystyle{ieeetr}

%

\vspace{-10.0 mm}
\begin{IEEEbiography}[{\includegraphics[width=1in,height=1.25in,clip,keepaspectratio]{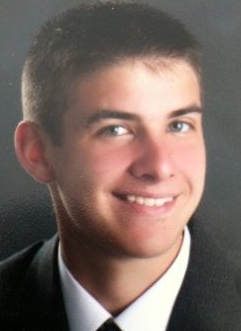}}]{Joshua J. Engelsma}
graduated magna cum laude with a B.S. degree
in computer science from Grand Valley State University, 
Allendale, Michigan, in 2016. He is currently
working towards a PhD degree in the
Department of Computer Science and Engineering
at Michigan State University, East Lansing,
Michigan. His research interests include pattern
recognition, computer vision, and image processing
with applications in biometrics.
\end{IEEEbiography}

\begin{IEEEbiography}[{\includegraphics[width=1in,height=1.25in,clip,keepaspectratio]{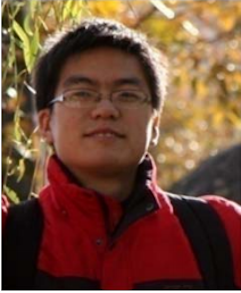}}]{Kai Cao}
received the Ph.D. degree from the
Key Laboratory of Complex Systems and Intelligence
Science, Institute of Automation, Chinese
Academy of Sciences, Beijing, China, in 2010.
He is currently a Post Doctoral Fellow in the Department
of Computer Science \& Engineering,
Michigan State University. He was affiliated with
Xidian University as an Associate Professor. His
research interests include biometric recognition,
image processing and machine learning.
\end{IEEEbiography}

\begin{IEEEbiography}[{\includegraphics[width=1in,height=1.25in,clip,keepaspectratio]{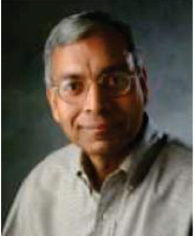}}]{Anil K. Jain}
is a University distinguished professor
in the Department of Computer Science
and Engineering at Michigan State University.
He was the editor-in-chief of the IEEE Trans.
Pattern Analysis and Machine Intelligence and
a member of the United States Defense Science
Board. He has received Fulbright, Guggenheim,
Alexander von Humboldt, and IAPR King
Sun Fu awards. He is a member of the National
Academy of Engineering and foreign fellow of the Indian National
Academy of Engineering.
\end{IEEEbiography}
\vfill




\end{document}